\documentclass[a4paper,11pt]{article}
\usepackage[utf8]{inputenc}
\usepackage{graphicx}
\usepackage{hyperref} 
\usepackage{amssymb}
\usepackage{amsmath}
\usepackage{xcolor}
\usepackage{bbm, dsfont}
\usepackage{multicol}
\usepackage{centernot}
\usepackage{overpic}
\usepackage{subcaption}
\usepackage{floatrow}
\usepackage{multirow}

\newcommand{\p}{\mathbb{P}}
\newcommand{\R}{\mathbb{R}}

   \newcommand\independent{\protect\mathpalette{\protect\independenT}{\perp}}
    \def\independenT#1#2{\mathrel{\rlap{$#1#2$}\mkern2mu{#1#2}}}

\def\argmin{\textit{argmin}}
\newtheorem {Proposition}{Proposition}[section]

\newtheorem {Theorem}[Proposition]{Theorem}

\newtheorem {Remark}[Proposition]{Remark}

\newtheorem {Assumption}[Proposition]{Assumption}

\numberwithin{equation}{section}

\title{Review of Mathematical frameworks for Fairness in Machine Learning}
\author{
 E. del Barrio$^2$ $\And$  P. Gordaliza$^{1,2}$ $\And$ J-M. Loubes$^1$  \\
      \\
  1\thanks{ We thank the AI interdisciplinary institute ANITI, grant agreement ANR-19-PI3A-0004 under the French investing for the future PIA3 program.} : Institut de Math\'ematiques de Toulouse\\
  Universit\'e de Toulouse 3\\
  Toulouse, France  \\
 \\
2: Instituto de Matemáticas de la Universidad de Valladolid\\
Dpto. de Estadistica e Investigacion Operativa \\
 Universidad de Valladolid\\
  Valladolid, Spain}

\begin{document}
	
	\maketitle

\begin{abstract}
A review of the main fairness definitions and fair learning methodologies proposed in the literature over the last years is presented from a mathematical point of view. Following our independence-based approach, we consider how to build fair algorithms and the consequences on the degradation of their performance compared to the possibly unfair case. This corresponds to the price for fairness given by the criteria \textit{statistical parity} or \textit{equality of odds}. Novel results giving the expressions of the optimal fair classifier and the optimal fair predictor (under a linear regression gaussian model) in the sense of \textit{equality of odds} are presented.

\end{abstract}

	\section{Introduction}
With both the introduction of new ways of storing, sharing and streaming data and the drastic development of the capacity of computers to handle large computations, the conception of models have changed. Mathematical models were first designed following prior ideas or conjectures from physical or biological models, then tested by designing experiments to test the validity of the ideas of their inventors. The model holds until new observations enable to reject its assumptions. The so-called Big Data's area introduced a new  paradigm. The observed data convey enough information to understand the complexity of real life and  the more the data, the better the description of the reality. Hence building models optimised to  fit the data  has become an efficient way to obtain generalizable models  able to describe and forecast  the real world.

In this framework, the principle of supervised machine learning is to build a decision rule from a set of labeled examples called the learning sample, that fits the data. This rule becomes a model or a decision algorithm that will be used for  all the population. Mathematical guarantees can be provided in certain cases to control  the generalization error of the algorithm which corresponds to the approximation done by building the model based on the observations and not knowing the true model that actually generated the data set. More precisely, the data are assumed to follow an unknown distribution while only its empirical distribution is at hand. So bounds are given to measure the error made by fitting a model on such observations and still using the model for new data. Yet the underlying assumption is that the observations follow all the same distribution which can be correctly estimated by the learning sample. Potential existing bias in the learning sample will be implicitly learnt and incorporated in the prediction. The danger of an uncontrolled prediction is greater when the algorithm lacks interpretability hence providing predictions that seem to be drawn from a yet accurate black-box but without any control or understanding on the reasons why they were chosen.

More precisely, in a supervised setting, the aim of a machine learning algorithm is to learn the relationships between characteristic variables $X$ and a target variable $Y$ in order to forecast new observations. Set the learning sample as  $(Y_1,X_1),\dots, (Y_n,X_n) $ i.i.d observations drawn from an unknown distribution $\mathbb{P}$. Set the empirical distribution $\mathbb{P}_n=\frac{1}{n} \sum_{i=1}^n \delta_{X_i,Y_i}$. The quality of the prediction will be measured using a loss function defined as $\ell  :  \: (Y,\hat{Y}) \mapsto \ell(Y,\hat{Y}) \in \mathbb{R}^+$ to quantify the error made while predicting $\hat{Y}$ when $Y$ is observed. Then for a given chosen class of algortihms $\mathcal{F}$, consider $\widehat{f}_n$ the best model that can be estimated by minimizing over  $\mathcal{F}$, the loss function (and possibly a penalty to prevent overfitting for example), namely 
\begin{equation}
\label{eq:main}	
\widehat{f}_n \in  {\rm arg}\min_{f \in \mathcal{F}}  \left\{  \frac{1}{n} \sum_{i=1}^n\ell (Y_i,f(X_i)) +  \lambda {\rm penalty}(f) \right\},
\end{equation}
where $\lambda$ balances the contribution of both terms to get a trade-off between the bias and the efficiency of the algorithm. The oracle rule is the best (yet unknown) rule that could be constructed if the true distribution were known $$ f^\star \in {\rm arg}\min_{f \in \mathcal{F}} \mathbb{E}_\mathbb{P}  \{ \ell (Y,f(X)) +  \lambda {\rm penalty}(f) \}. $$ The predictions are given by $\widehat{Y} = \widehat{f}_n (X).$ Results from machine learning theory ensures that for proper choices of set of rules $\mathcal{F}$, the prediction's error behaves close to the oracle in the sense that, from a mathematical point of view, the excess risk
$$  \mathbb{E}_\mathbb{P}  \{ \ell (Y, \widehat{f}_n (X) )  \} - \mathbb{E}_\mathbb{P}  \{ \ell (Y, {f}^\star (X) )  \} $$ is small. So mathematical guarantees warrant that the optimal forecast model reproduces the uses learnt from the learning set for new observations. It shapes the reality according to the learnt rule without questioning nor evolution. 

\section{A definition of fairness in machine learning as independence criterion} \label{s:definition}
\subsection{Definition of full fairness} \label{subs:PFairness}
There is no doubt that machine learning is a powerful tool that is improving human life and has shown great promise in the developping of very different technological applications, including powering self-driving cars, accurately recognizing cancer in radiographs, or predicting our interests based upon past behavior, to name just a few. Yet with its benefits, machine learning also involves delicate issues such as the presence of bias in the model classifications and predictions. Hence, with this generalization of predictive algorithms in a wide variety of fields, algorithmic fairness is gaining more and more attention not only in the research community but also among the general population, who is experiencing a great impact on its daily life and activity. Thanks to this, there has been a push for the emergence of different approaches for assessing the presence of bias in machine learning algorithms over the last years. Similarly, various classifications have been proposed to understand the different sources of data bias. We refer to \cite{mehrabi2019survey} for a recent review.

\begin{figure}[ht]
	\ffigbox{
	\begin{subfloatrow}
		\ffigbox{
		\includegraphics[width=.8\linewidth]{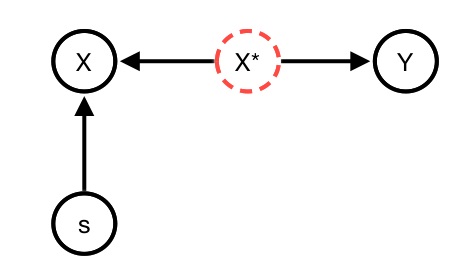} 
	}{
		\caption{}
		\label{fig:Bmodel1}
	}
\ffigbox{

		\includegraphics[width=.8\linewidth]{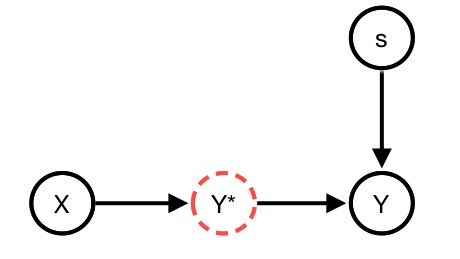} 
	}{
		\caption{}
		\label{fig:Bmodel2}
}
	\end{subfloatrow}
}{
	\caption{Two models for understanding the introduction of bias in the model}
	\label{fig:Bmodels}
}
\end{figure}

Consider the probability space $\left(\Omega\subset \R^d, \mathcal{B}, \p\right)$, with $\mathcal{B}$ the Borel $\sigma-$algebra of subsets of $\R^d$ and $d \geq 1$. We will assume in the following that the bias is modeled by the random variable $S \in \mathcal{S}$ that represents an information about the observations $X \in \mathcal{X} \subset \R^d,$ that should not be included in the model for the prediction of the target $Y \in \R^d, \ d \geq 1$. In the fair learning literature, the variable $S$ is referred to as the \textit{protected} or \textit{sensitive attribute}. We assume moreover that this variable is observed. Most fairness theory has been developed particularly in the case when $\mathcal{S}=\{0,1\}$ and $S$ is a sensitive binary variable. In other words, the population is supposed to be possibly divided into two categories, taking the value $S=0$ for the \textit{minority} (assumed to be the unfavored class), and $S=1$ for the \textit{default} (and usually favored class). Hence, we also study more deeply this case and it will be conveniently indicated in the rest of the chapter, but in principle we consider general $\mathcal{S}$. From a mathematical point of view, we follow recent paper \cite{serrurier2019fairness} that proposed the two following models that aim at understanding how this bias could be introduced in the algorithms:
\begin{enumerate}
	\item The first model corresponds to the case where the data are subject to a bias nuisance variable which, in principle, is assumed not to be involved in the learning task, and whose influence in the prediction should be removed. We refer here to the well-known example of the dog vs. wolf in \cite{ribeiro2016should}, where the input data were images highly biased by the presence of background snow in the pictures of wolves, and the absence of it in those of dogs. As shown in Figure \ref{fig:Bmodel1}, this situation appears when the attributes $X$ are a biased version of unobserved fair attributes $X^\star$ and the target variable $Y$ depends only on $X^\star$. In this framework, learning from $X$ induces biases while fairness requires:
		\begin{equation*}
	X^\star \independent S\mid Y \quad \text{and} \quad Y \independent S\mid X^\star.
	\end{equation*}
	Note that either $X$ nor $Y$ is independent of the protected $S$.
	
	\item The second model corresponds to the situation when a biased decision is observed as a result of a fair score $Y^\star$ which has been biased by the uses giving rise to the target $Y$. Thus, a fair model in this case will change the prediction in order to make them independent of the protected variable. This is represented in Figure \ref{fig:Bmodel2} and, formally, it is required that 
	\begin{equation*}
	X \independent S\mid Y \quad \text{and} \quad Y^\star \independent S\mid Y,
	\end{equation*}
	where $Y^\star$ is not observed. Note that previous conditions do not imply the independence between $Y$ and $S$ (even conditionally to $X$).

\end{enumerate}

In the statistical literature, an algorithm $\hat{f}_n$ is called fair or unbiased when its outcome does not depend on the sensitive variable. The notion of \textit{perfect fairness} requires that the protected variable $S$ does not play any role in the forecast $\widehat{Y}=f(X,S)$ of the target $Y$. In other words, we will be looking at the independence between the protected variable $S$ and the outcome $\widehat{Y}$, both considering given or not the true value of the target $Y$. These two notions of fairness are known in the literature as:
\begin{itemize}
	\item \textit{Statistical parity} (S.P.) deals with the independence between the outcome of the algorithm and the sensitive attribute 
	\begin{equation}\label{SPindependence}
	\hat{Y} \independent S
	\end{equation}
	\item \textit{Equality of odds} (E.O.) considers the independence between the protected attribute and the outcome conditionally given the true value of the target
	\begin{equation}\label{EOindependence}
\hat{Y}  \independent S \: | \: Y 
	\end{equation}
\end{itemize}
Hence, a perfect fair model should be chosen within a class ensuring one of these restrictions \eqref{SPindependence}-\eqref{EOindependence}. Observe that the choice of the notion of fairness is convenient regarding the assumed model for the introduction of the bias in the algorithm: while \textit{statistical parity} is suitable for model \ref{fig:Bmodel1}, \textit{equality of odds} is for model \ref{fig:Bmodel2}, and especially well-suited for scenarios where ground truth is available for historical decisions used during the training phase. 

In this work, we tackle only these two main notions of fairness developed among the machine learning community. There are other definitions such as \textit{avoiding disparate treatment} or \textit{predictive parity}, defined respectively as $ \hat{Y} \: | \: X  \independent S$ or $ Y   \independent S \: | \: \hat{Y}$. A decision making system suffers from \textit{disparate treatment} if it provides different outcomes for different groups of people with the same (or similar) values of non-sensitive features but different values of sensitive features \cite{barocas2016big}. In other words, (partly) basing the decision outcomes on the sensitive feature value amounts to disparate treatment. Technically, the \textit{disparate treatment} doctrine tries to counter explicit as well as intentional discrimination \cite{barocas2016big}. It follows from the specification of \textit{disparate treatment} that a decision maker with an intent to discriminate could try to disadvantage a group with a certain sensitive feature value (e.g., a specific race group) not by explicitly using the sensitive feature itself, but by intentionally basing decisions on a correlated feature (e.g., the non-sensitive feature location might be correlated with the sensitive feature race). This practice is often referred to as redlining in the US anti-discrimination law and also qualifies as disparate treatment \cite{gano2017disparate}. However, such hidden intentional disparate treatment maybe be hard to detect, and some authors argue that \textit{statistical parity} might be a more suitable framework for detecting such covert discrimination \cite{siegel2014race}, while others focus only on explicit disparate treatment \cite{zafar2019fairness}. For further details, we refer to the comprehensive study of fairness in machine learning given in \cite{barocas-hardt-narayanan}.

The description of the metrics given above applies in a general context, yet all four fairness measures were originally proposed within the binary classification framework. Hence the literature cites and equivalent denominations will be presented in the following subsection specifically for this context. 

\subsection{The special case of  classification}
Fairness has been widely studied in the binary classification setting. Here the problem consists in forecasting a binary variable $Y \in \left\lbrace0,1\right\rbrace$, using observed covariates  $X \in \R^d, \ d\geq 1.$ We introduce also a notion of positive prediction: $Y=1$ represents a \textit{success} while $Y=0$ is a \textit{failure}. We refer to \cite{bousquet2004introduction} for a complete description of classification problems in statistical learning. In this framework, the two main algorithmic fairness metrics are specified as follows.

\begin{itemize}
	\item \textit{Statistical parity.} Despite the early uses of this notion through the so-called $4/5^{th}$-rule for fair classification purposes by the State of California Fair Employment Practice Commission (FEPC) in 1971\footnote{https://www.govinfo.gov/content/pkg/CFR-2017-title29-vol4/xml/CFR-2017-title29-vol4-part1607.xml}, it was first formally introduced as \textit{statistical parity} in \cite{dwork2012fairness} in the particular case when $S$ is also binary. Since then it has received several other denominations in the fair learning literature. For instance, it has been equivalently named in the same introductory work as \textit{demographic parity} or \textit{group fairness}, and also in others \textit{equal acceptance rate} \cite{zliobaite2015relation} or \textit{benchmarking} \cite{simoiu2017problem}. Formally, if $S \in \{0,1\}$ this definition of fairness is satisfied when both subgroups are equally probable to have a successful outcome
	\begin{equation}\label{def:StatisticalParity}
	\mathbb{P}(\hat{Y}=1 \mid S=0)= \mathbb{P}(\hat{Y}=1 \mid S=1),
	\end{equation}
	which can be extended to $\mathbb{P}(\hat{Y}=1 \mid S)= \mathbb{P}(\hat{Y}=1)$ for general $S$, continuous or discrete. A related and more rigid measure is called \textit{avoiding disparate treatment} in \cite{ZVGG} if the probability that the classifier outputs a specific value of the forecast given a feature vector does not change after
	observing the sensitive feature, namely $\mathbb{P}(\hat{Y}=1 \mid X, S)= \mathbb{P}(\hat{Y}=1 \mid X)$.
	\item \textit{Equality of odds} (or \textit{equalized odds}) looks for the independence between the error of the algorithm and the protected variable. Hence, in practice, when $S$ is also binary it compares the error rates of the algorithmic decisions between the different groups of the population, and considers that a classifier is fair when both classes have equal False and True Positive Rates
	\begin{equation}\label{def:EqOdds}
	\mathbb{P}(\hat{Y}=1\mid Y=i, S=0)= \mathbb{P}(\hat{Y}=1\mid Y=i, S=1), \ \text{for} \ i=0,1.
	\end{equation}
	For general $S$, we note that this condition is equivalent to
	\begin{equation}
	\mathbb{P}(\hat{Y}=1\mid Y=i, S)= \mathbb{P}(\hat{Y}=1\mid Y=i), \ \text{for} \ i=0,1.
	\end{equation}
	This second point of view was introduced in \cite{hardt2016equality} and has been originally proposed for recidivism of defendants in \cite{flores2016false}. Over the last few years it has been given several names, including \textit{error rate balance} in \cite{chouldechova2017fair} or \textit{conditional procedure accuracy equality} in \cite{berk2018fairness}.
\end{itemize}
Many other metrics have received significant recent attention in the classification literature. In this setting, the already cited above \textit{disparate treatment}, also referred to as \textit{direct discrimination} \cite{pedreshi2008discrimination}, looks at the equality for all $x \in \mathcal{X}$
\begin{equation}
	\mathbb{P}(\hat{Y}=1\mid X=x, S=0)= \mathbb{P}(\hat{Y}=1\mid X=x, S=1)
\end{equation}
 Furthermore, we note that \textit{equality of opportunity} (\cite{hardt2016equality} or \cite{kusner2017counterfactual}) and \textit{avoiding disparate mistreatment} \cite{ZVGG} are two metrics related to the previous \textit{equalized odds}, yet weaker. The first one requires only the equality of true positive rates, that is when $i=1$ in \eqref{def:EqOdds}, while the second looks at the equality of misclassification error rates across the groups:
\begin{equation}\label{def:DM}
\mathbb{P}(\hat{Y}\neq Y\mid S=0)= \mathbb{P}(\hat{Y}\neq Y\mid S=1).
\end{equation} 
Thus, \textit{equality of odds} implies both the lack of \textit{disparate mistreatment} and \textit{equality of opportunity}, but not viceversa. Finally, we mention also here \textit{predictive parity} which was introduced in \cite{chouldechova2017fair}. It requires the equality of positive predictive values across both groups. Therefore, mathematically it is satisfied when 
\begin{equation}\label{def:predictiveparity}
\mathbb{P}(Y=1\mid \hat{Y}=1, S=0)= \mathbb{P}(Y=1\mid \hat{Y}=1, S=1).
\end{equation}

The fairness metrics defined above are evaluated only for binary predictions and outcomes. By contrast, we can find also in the literature a set of metrics involving explicit generation of a continuous-valued score denoted here by $R \in [0,1]$. Although scores could be used directly, they can alternatively serve as the input to a thresholding function that outputs a binary prediction.

Among this set, we highlight the notion of \textit{test-fairness}, which extends \textit{predictive parity} \eqref{def:predictiveparity} when the prediction is a score. An algorithm satisfies this kind of fairness (or it is said to be \textit{calibrated}) if for all scores $r$, the individuals who have the same score have the same probability of belonging to the positive class, regardless of group membership. Formally, this is expressed as $\mathbb{P}(Y=1\mid R=r, S=0)= \mathbb{P}(Y=1\mid R=r, S=1),$ for all scores $r$. This criteria was introduced in \cite{chouldechova2017fair} and has also been termed as \textit{matching conditional frequencies} by \cite{hardt2016equality}.

A related metric called \textit{well-calibration} \cite{verma2018fairness} or \textit{calibration within groups} \cite{kleinberg2016inherent} imposes an additional and
more stringent condition: a model is \textit{well-calibrated} if individuals
assigned score $r$ must have probability exactly $r$ of belonging to the positive class. If this condition is satisfied, then \textit{test-fairness}
will also hold automatically, though not viceversa. Indeed, we note that the scores of a calibrated predictor can be transformed into scores satisfying well-calibration.

Finally, \textit{balance for positive/negative class} was introduced in \cite{kleinberg2016inherent} as a generalization of the notion of \textit{equality of odds}. Mathematically, this balance is expressed through the equalities of expected values
$\mathbb{E}(R \mid Y=i, S=0)=\mathbb{E}(R \mid Y=i, S=1), \ i\in\{0,1\}$.

\subsection{Relationships between fairness criteria}\label{subsec:relation}
It is also important to note that the wide variety of the proposed criteria formalizing different notions of fairness (see reviews \cite{berk2018fairness} and \cite{verma2018fairness} for more details) has lead sometimes to incompatible formulations. The conditions under which more than one metric can be simultaneously satisfied, and relatedly, the ways in which different metrics might be in tension have been studied in several works \cite{chouldechova2017fair, kleinberg2016inherent, berk2018fairness}. Indeed, in the following Propositions \ref{prop:SPvsEO}, \ref{prop:SPvsPP}, \ref{prop:PPvsEO} we revisit three \textit{impossibility theorems of fairness} stating the exclusivity, except in non-degenerate cases, of the three main criteria considered in fair learning.

We study first the combination of all three of these metrics and then explore conditions under which it may be possible to simultaneously satisfy two metrics. To begin with, it is interesting to note that from the definition of conditional probability, the respective probability distributions associated with each of these three fairness metrics can be expressed as follows:
\begin{align}
\mathcal{L}(Y,\hat{Y} \mid S)&=\mathcal{L}(Y\mid \hat{Y}, S) \times \mathcal{L}(\hat{Y}\mid S)\label{eq:compFM1}\\
&= \mathcal{L}(\hat{Y}\mid Y, S) \times \mathcal{L}(Y\mid S)\label{eq:compFM2}.
\end{align}
We observe that on the right-hand side of equality \eqref{eq:compFM1} the first factor refers to \textit{predictive parity}, while the second one to \textit{statistical parity}. Similarly, in the equality \eqref{eq:compFM2} the first term represents \textit{equality of odds} while the second one the base rate, that is the distribution of the true target among each group.

While the three results for fairness incompatibilities are stated hereafter in a general learning setting and their proofs are gathered in the Appendix \ref{app:relation}, in this section we present a discussion in the binary classification framework. Let us consider then the following notations for $s\in\{0,1\},$
\begin{itemize}
	\item $TPR_s:=\mathbb{P}(\hat{Y}=1 \mid Y=1, S=s)$ the group-specific true positive rates
	\item $FPR_s:=\mathbb{P}(\hat{Y}=1 \mid Y=0, S=s)$ the group-specific false positive rates
	\item $PPV_s:=\mathbb{P}(Y=1 \mid \hat{Y}=1 , S=s)$ the group-specific positive predictive values
\end{itemize}

We consider first if a predictor can simultaneously satisfy \textit{equalized odds} and \textit{statistical parity}.
\begin{Proposition}[Statistical parity vs. Equality of odds]\label{prop:SPvsEO}
If $S$ is dependent of $Y$ and $\hat{Y}$ is dependent of $Y$, then either statistical parity holds or equality of odds but not both.
\end{Proposition}

In the special case of binary classification the result can be sharpened as follows. Observe that we can write for $s \in \{0,1\},$
\begin{align}\label{eq:successcond}
\mathbb{P}(\hat{Y}=1 \mid S=s)=\mathbb{P}(Y=1 \mid S=s)TPR_s+\mathbb{P}(Y=0 \mid S=s)FPR_s
\end{align}
Then computing the difference between expression \eqref{eq:successcond} for each class and assuming that \textit{equalized odds} holds, namely $$TPR_0=TPR_1=\mathbb{P}(\hat{Y}=1 \mid Y=1) \ \text{and} \  FPR_0=FPR_1=\mathbb{P}(\hat{Y}=1 \mid Y=0),$$ we obtain
\begin{align*}
&\mathbb{P}(\hat{Y}=1 \mid S=0)-\mathbb{P}(\hat{Y}=1 \mid S=1)\\
&=(\mathbb{P}(Y=1 \mid S=0)-\mathbb{P}(Y=1 \mid S=1))\mathbb{P}(\hat{Y}=1 \mid Y=1)\\
&+(\mathbb{P}(Y=0 \mid S=0)-\mathbb{P}(Y=0 \mid S=1))\mathbb{P}(\hat{Y}=1 \mid Y=0)\\
&=(\mathbb{P}(Y=1 \mid S=0)-\mathbb{P}(Y=1 \mid S=1))(\mathbb{P}(\hat{Y}=1 \mid Y=1)-\mathbb{P}(\hat{Y}=1 \mid Y=0))
\end{align*}
\textit{Statistical parity} requires that left side is exactly zero. Hence, for the right side also being zero necessarily $\mathbb{P}(Y=1 \mid S=0)=\mathbb{P}(Y=1 \mid S=1)$ or $\mathbb{P}(\hat{Y}=1 \mid Y=1)=\mathbb{P}(\hat{Y}=1 \mid Y=0)$. However, it is usually assumed that base rates differs across the groups, that is, the ratio of people in the group who belong to the positive class ($Y=1$) to the total number of people in that group. Thus, \textit{statistical parity} and \textit{equalized odds} are simultaneously achieved only if true and false positive rates are equal. While this is mathematically possible, such condition is not particularly useful since the goal is typically to develop a predictor in which the true positive rate is significantly higher than the false.

\begin{Proposition}[Statistical parity vs. Predictive parity]\label{prop:SPvsPP}
	If $S$ is dependent of $Y$, then either statistical parity holds or predictive parity but not both.
\end{Proposition}

By contrast, in the binary classification setup the two fairness metrics are actually simultaneously feasible. Assume that \textit{statistical parity} holds, that is, $\mathbb{P}(\hat{Y}=1| S=1)=\mathbb{P}(\hat{Y}=1|S=0)=\mathbb{P}(\hat{Y}=1)$. Then, from equations \eqref{eq:compFM1}-\eqref{eq:compFM2} we can write the difference of positive predictive values
\begin{align}
PPV_0-PPV_1=\frac{TPR_0 \mathbb{P}(Y=1| S=0) - TPR_1 \mathbb{P}(Y=1| S=1)}{\mathbb{P}(\hat{Y}=1)}
\end{align}
Under \textit{predictive parity} the left side of the above equation must be zero, which in turn requires that the ratio of the true positive rates of the two groups be the reciprocal of the ratio of the base rates, namely
\begin{align}
\frac{TPR_0}{TPR_1}=\frac{\mathbb{P}(Y=1| S=1)}{\mathbb{P}(Y=1| S=0)}
\end{align}
Thus, while statistical and predictive parity can be simultaneously satisfied even with different base rates, the utility of such a predictor is limited when the ratio of the base rates differs significantly from 1, as this forces the true positive
rate for one of the groups to be very low.

\begin{Proposition}[Predictive parity vs. Equality of odds]\label{prop:PPvsEO}
	If $S$ is dependent of $Y$ then either predictive parity holds or equality of odds but not both.
\end{Proposition}

We explore this incompatibility in more detail in the binary classification framework. If both conditions hold
\begin{equation}\label{eq:Requalities}
TPR_0=TPR_1, \ \ FPR_0=FPR_1, \ \ \text{and} \ \ PPV_0=PPV_1,
\end{equation}
so we can write
\begin{align*}
\mathbb{P}(\hat{Y}=1 \mid S)=\sum_{i=0,1}\mathbb{P}(\hat{Y}=1 \mid Y=i, S)\mathbb{P}(\hat{Y}=1 \mid S)=TPR_0\mathbb{P}(Y=1 \mid S)+FPR_0\mathbb{P}(Y=0 \mid S).
\end{align*}
This together with equations \eqref{eq:compFM1}-\eqref{eq:compFM2} implies
\begin{align*}
&\mathbb{P}(\hat{Y}=1|Y=1, S=0)\mathbb{P}(Y=1| S=0)\\
&=\mathbb{P}(y=1| \hat{Y}=1, S=0)\left[TPR_0\mathbb{P}(\hat{Y}=1| S)+FPR_0\mathbb{P}(Y=0|S=0)\right],
\end{align*}
and using the notations above we obtain
\begin{align*}
TPR_0\mathbb{P}(Y=1| S=0)=PPV_0\left[TPR_0\mathbb{P}(\hat{Y}=1| S)+FPR_0(1-\mathbb{P}(Y=1|S=0))\right].
\end{align*}
Finally, we obtain the following expressions for the group-specific base rate for $s=0$
\begin{align}\label{eq:BR0}
\mathbb{P}(Y=1| S=0)=\frac{PPV_0FPR_0}{PPV_0FPR_0+(1-PPV_0)TPR_0}
\end{align}
and reasoning likewise for $s=1$
\begin{align}\label{eq:BR1}
\mathbb{P}(Y=1| S=1)=\frac{PPV_1FPR_1}{PPV_1FPR_1+(1-PPV_1)TPR_1}
\end{align}

 Hence, in the absence of perfect prediction, under assumption \eqref{eq:Requalities} base rates have to be equal for both \textit{equalized odds} and \textit{predictive parity} to simultaneously hold. When perfect
prediction is achieved, equations \eqref{eq:BR0} and \eqref{eq:BR1} take on the indefinite form $0/0$ so therefore do not convey anything definitive about base rates in that scenario.

We also note that the less strict metric \textit{equal opportunity} (recall it requires only equal TPR across groups) is compatible with \textit{predictive parity}. This is evident from equations \eqref{eq:BR0} and \eqref{eq:BR1} when the condition $FPR_0 = FPR_1$ is removed, thereby allowing \textit{equalized opportunity} and \textit{predictive parity} to be simultaneously satisfied even with unequal base rates. However, achieving this condition with unequal base rates will require that the FPR differs across the groups. When the difference between the base rates is large, the variation between group-specific FPRs may have to be significant which may reduce suitability for some applications. Hence, while \textit{equal opportunity} and \textit{predictive parity} are compatible in the presence of unequal base rates, practitioners should consider the cost (in
terms of FPR difference) before attempting to simultaneously achieve both. A similar analysis is possible when we considering parity in negative predictive value instead of positive predictive value, i.e. equal opportunity and parity in
NPV are compatible, but only at the cost of variation between group-specific true negative rates (TNRs).

\section{Price for fairness in machine learning}\label{s:price}

In this section, we consider how to build fair algorithms and the consequences on the degradation of their performance compared to the {\it possibly unfair} case. This corresponds to the price for fairness. \vskip .1in
Recall that the performance of an algorithm is measured through its risk defined by
$$R (f) = \mathbb{E} (\ell(Y,f(X,S))).$$
Define some class or restriction of classes
\begin{align}
&\mathcal{F}_{SP}=\{f(X,S) \in \mathcal{F} \quad {\rm s.t} \quad  \hat{Y} \independent S \} \label{defi:FairClassSP} \\
&\mathcal{F}_{EO}=\{f(X,S) \in \mathcal{F} \quad {\rm s.t} \quad \hat{Y} |Y  \independent S\} \label{defi:FairClassEO}
\end{align}
From a theoretical point of view, a fair model can be achieved by restricting the minimization \eqref{eq:main} to such classes.
The price for fairness is 
\begin{equation} \label{eq:theprice}	
\mathcal{E}_{\rm Fair}(\mathcal{F}):=\inf_{f \in \mathcal{F}_{\rm Fair}} R(f) - \inf_{f \in \mathcal{F}} R(f).
\end{equation}
If $\mathcal{F}$ denotes the class of all measurable functions, then $\inf_{f \in \mathcal{F}} R(f)$ is known as the Bayes Risk. In the following, we will study the difference of the minimal risks in \eqref{eq:theprice} under both fairness assumptions and in two different frameworks: regression and classification.

To address this issue, we will consider the  Wasserstein (a.k.a Monge-Kantorovich) distance between distributions. The Wasserstein distance appears as an appropriate tool for comparing probability distributions and arises naturally from the optimal transport problem (we refer to \cite{villani2008optimal} for a detailed description). For $P$ and $Q$ two probability measures on $\mathbb{R}^d$, the squared Wasserstein distance between $P$ and $Q$ is defined as
\[
W_2^2(P,Q):=\min_{\pi \in\Pi(P,Q)} \int \|x-y\|^2 d\pi(x,y)
\]
where
$\Pi(P,Q)$ the set of  probability measures on $\mathbb{R} ^d\times \mathbb{R}^d$ with marginals $P$ and $Q$.

\subsection{Price for fairness as Statistical Parity}

The notion of perfect fairness given by \textit{statistical parity} criterion implies that the distribution of the predictor does not depend on the protected variable $S$.

\subsubsection{Regression}\label{subsubsec:PFSPregression}
In the regression problem, \textit{statistical parity} condition is expressed through the equality of distributions $\mathcal{L}(f(X,S)|S)= \mathcal{L}(f(X,S))$.
Then in this setting a standard definition of this statistical independence requires that $\mathbb{P}(f(X,s) \in A | S=s) = \mathbb{P}(f(X,S) \in A)$ for all $s \in \mathcal{S}$ and all measurable sets $A$. Since $f(X,S)$ is a real-valued random variable under Borel $\sigma$-algebra, it is fully characterized by its cumulative distribution function, and so it suffices to consider sets $A= [0, +\infty)$.

This fairness assumption implies the weakest cases where $\mathbb{E}(f(X,S)|S)= \mathbb{E}(f(X,S)) $ as presented in \cite{dwork2012fairness, zemel2013learning}, or equivalently when ${\rm Cov}(f(X,S),S)=0$. Note that in the case where $S$ is a discrete variable, previous criteria have a simpler expression.
In particular, in the binary setup when $S \in \{0,1\}$, we can write
\begin{align*}
\mathbb{E}_{X,S}(f(X,S))&=\mathbb{E}_X[\mathbb{E}_S[f(X,S)\mid S]]\\
&=\mathbb{P}(S=0)\mathbb{E}_X(f(X,S)\mid S=0)+\mathbb{P}(S=1)\mathbb{E}_X(f(X,S)\mid S=1).
\end{align*}
On the other hand, the definition of conditional expectation gives
\begin{align*}
\mathbb{E}_{X,S}(f(X,S)\mid S)&=\frac{\mathbb{E}_{X,S}[Sf(X,S)]}{\mathbb{E}(S)}
=\frac{\mathbb{P}(S=1)\mathbb{E}_X(f(X,S)\mid S=1)}{\mathbb{P}(S=1)}\\
&=\mathbb{E}_X(f(X,S)\mid S=1).
\end{align*}
From both equalities above we have that \textit{statistical parity} holds if and only if
\begin{align*}
\mathbb{P}(S=0)\mathbb{E}_X(f(X,S)\mid S=0)+\mathbb{P}(S=1)\mathbb{E}_X(f(X,S)\mid S=1)=\mathbb{E}_X(f(X,S)\mid S=1),
\end{align*}
which, if $P(S=0)>0$, reduces to 
\begin{align*}
\mathbb{E}_X(f(X,S)\mid S=0)=\mathbb{E}_X(f(X,S)\mid S=1).
\end{align*}

In the general regression setting, we will use the following notations : $X\in \mathcal{X}$, $S \in \mathcal{S}$, $Y \in \mathbb{R}^d$
When $\mathcal{F}$ is the set of all measurable functions from $\mathcal{X} \times \mathcal{S}$ to $\mathbb{R}^d$, the optimal risk (a.k.a. Bayesian risk), is defined as 
\[
R^\star:=\mathcal{R}(\mathcal{F})=\min_{f \in \mathcal{F}} \mathbb{E} \|Y-f(X,S)\|^2
\]
is achieved for the Bayes estimator $$ \eta(X,S):=\mathbb{E} [Y|(X,S)]. $$ 
Denote  $\mu_S$ the conditional distribution of the Bayes estimator $\mathbb{E}(Y|X,S)$ given $S$ and for a predictor $g$  $\nu_S(g)$ the conditional distribution of $g(X,S)$ given $S$. In \cite{legouic2020fair} the authors relate the excess risk with a minimization problem in the Wasserstein space proving the following lower bound for the price for fairness.
\begin{Theorem}
	\begin{equation}\label{eq:bound}
	\inf_{f \in \mathcal{F}_{\rm Fair}} R(f) - \inf_f R(f)  \geq \inf_{g  \in \mathcal{F}} \mathbb{E} W_2^2(\mu_S,\nu_S(g)). 
	\end{equation}
	Moreover, if $\mathcal{F}=\mathcal{F}_{SP}$  and $\mu_s$ has density w.r.t. Lebesgue measure for almost every $s$, then \eqref{eq:bound} becomes an equality
	\begin{equation}
	\label{eq:fairreg}
	\mathcal{E}_{\rm Fair}(\mathcal{F}) = \inf_{g \in \mathcal{F}} \mathbb{E}_S W_2^2(\mu_S,\nu_S(g)).
	\end{equation}
\end{Theorem}
Imposing fairness comes at a price that can be quantified which depends on the 2-Wasserstein distance between distributions of Bayes predictors.

Finding the minimum in \eqref{eq:fairreg} is related to the minimization of  Wasserstein's variation which has been known as the problem of studying Wasserstein's barycenter.
Actually, for Statistical Parity constraint 
$$\inf_{g \in \mathcal{F}} \mathbb{E}_S W_2^2(\mu_S,\nu_S(g))=\inf_{\nu(g) } \mathbb{E}_S W_2^2(\mu_S,\nu(g)) $$
which  amounts to  minimize $$ \nu \mapsto \mathbb{E}_S W_2^2(\mu_S,\nu) $$ 

This problem has been studied in \cite{agueh2011barycenters}, \cite{le2017existence} or \cite{del2019central}. The distributions $\mu_S$ are random distributions and define $\mathbb{P}_{\cal{S}}$ their distribution on the set of distributions. Hence
The minimum is reached for $\mu_B$ the Wasserstein barycenter of $\mathbb{P}_{\cal{S}}$. Note that if  $S$ is discrete, in particular for the two class version $S \in \{0,1\}$, note $\pi_s=P(S=s)$, the distribution $\mathbb{P}_{\cal{S}}$ can be written as $\mathbb{P}_{\cal{S}}= \pi_1 \delta_{\mu_1}+(1-\pi_1) \delta_{\mu_0}$. Hence its barycenter is a measure that minimizes the functional 
$$ \nu \mapsto  \pi_0 W_2^2(\mu_0,\nu)+ (1-\pi_0) W_2^2(\mu_1,\nu). $$ Existence and uniqueness are ensured as soon as the $\mu_S$ have density with respect to Lebesgue measure.

\subsubsection{Classification}
We consider the problem of quantifying the price for imposing \textit{statistical parity} when the goal is predicting a label. In the following and without loss of generality, we assume that $Y$ is a binary variable with values in $\{0,1\}$. If $S$ is also binary, then Statistical Parity is often quantified in the fair learning literature using the so-called Disparate Impact (DI)
\begin{equation}\label{def:DIclassifier}
DI(g,X,S)=\frac{\mathbb{P}(g(X,S)=1 \mid S=0)}{\mathbb{P}(g(X,S)=1 \mid S=1)}.
\end{equation} 
This measures the risk of discrimination when using the decision rule encoded in $g$ on data following the same distribution as in the test set. Hence, in \cite{gordaliza2019obtaining} a classifier $g$ is said not to have a Disparate Impact at level $\tau \in \left(0,1\right]$ when $DI(g,X,S) > \tau$.  Perfect fairness is thus equivalent to the assumption that the disparate impact is exactly $ DI(g,X,S)=1.$ Note that the notion of DI defined Eq.~\eqref{def:DIclassifier} was first introduced as the $4/5^{th}$-rule by the State of California Fair Employment Practice Commission (FEPC) in 1971. Since then, the threshold $\tau_0=0.8$ was chosen in different trials as a legal score to judge whether the discriminations committed by an algorithm are acceptable or not (see e.g. \cite{FFMSV} \cite{ZVGG}, or \cite{mercat2016discrimination}).

While in the classification problem the notion of \textit{statistical parity} can be easily extended for general $S \in \mathcal{S}$, continuous or discrete, through the equality $\mathbb{P}(g(X,S)=1)=\mathbb{P}(g(X,S)=1 \mid S)$, the index Disparate Impact has not been used in the literature for quantifying fairness in the general framework. Hence, we only consider the classification problem. Still, if $S$ is a multiclass sensitive variable, we observe that a fair classifier should satisfy for all $s \in \mathcal{S}$,
\begin{equation}
\mathbb{P}(g(X,S)=1)=\mathbb{P}(g(X,S)=1 \mid S=s).
\end{equation}
Hence, Disparate Impact could be extended to
\begin{equation}\label{def:DIclassifierSmulticlass}
DI(g,X,S)=\frac{\min_{s \in \mathcal{S}}{\mathbb{P}(g(X,S)=1 \mid S=s)}}{\mathbb{P}(g(X,S)=1 \mid S=1)}.
\end{equation}
Tackling the issue of computing a bound in \eqref{eq:theprice} is a difficult task and has been studied by several authors. In this specific framework, finding a lower bound for the loss of accuracy induced by the full statistical parity constraint has not been solved. This is mainly due to the fact that the classification setting does not specify a model to constrain the relationships between the labels $Y$ and the observations $X$, enabling a too large choice of models, contrary to the regression case.

Yet in different frameworks, some results can be proved. On the one hand, in \cite{jiang2019wasserstein} a notion of fairness is considered which correspond to  controling the number of class changes when switching labels, which amounts to study the difference between classification errors for plug in rules corresponding to all possible thresholds $\tau$ of Bayes score called the model belief, $\eta_S(X)=P(Y=1|X,S) \geq \tau$. Authors achieve a bound using  the $W_1$ distance and prove that the minimum loss is achieved for the 1-Wasserstein barycenter.

In the following we recall results obtained in \cite{gordaliza2019obtaining} which study the price for fairness in statistical parity in the framework where we want to ensure that all classifiers trained by a transformation of the data will be fair with respect to the statistical parity definition.

For this consider the Balanced Error Rate $$BER(g,X,S)=\frac{\mathbb{P}\left(g(X,S)=0 \mid S=1\right)+\mathbb{P}\left(g(X,S)=1 \mid S=0\right)}{2}$$
corresponding to the problem of estimating the sensitive label  from the prediction in the most difficult case where the class are well balanced between each group labeled by the variable $S$. In this setting, unpredictability of the label  warrants the fairness of the procedure. Actually, given $\varepsilon >0, S$ is not $\varepsilon-$predictable from $X$ if $BER(g,X,S) > \varepsilon$, for all $g \in \mathcal{G}$

\begin{equation*}
DI(g,X,S):= \frac{a(g)}{b(g)}.
\end{equation*}
We consider classifiers $g$ such that $a(g)>0$ and $b(g)>0$.
\begin{Theorem}[Link between Disparate Impact and Predictability]
	Given random variables $X: \Omega \rightarrow \R^d, \ S: \Omega \rightarrow\{0,1\}$, the classifier $g \in \mathcal{G}$ has Disparate Impact at level $\tau \in \left[0,1\right]$, with respect to $(X,S)$, if, and only if, $S$ is $\left(\frac{1}{2}-\frac{a(g)}{2}(\frac{1}{\tau}-1)\right)-$predictable from $X$.
\end{Theorem}

Then, we can see that the notion of predictability and the distance in Total Variation between the conditional distributions of $X \mid S$ are connected through the following theorem

\begin{Theorem}[Total Variation distance]
	Given the variables $X: \Omega \rightarrow \R^d, \ d \geqslant 1,$ and $S: \Omega \rightarrow \{0,1\}$,
	\begin{equation*}
	\min_{g \in \mathcal{F}}BER(g,X,S)=\frac{1}{2}\left(1-d_{TV}\left(\mathcal{L}\left(X|S=0\right), \mathcal{L}\left(X|S=1\right)\right)\right),
	\end{equation*}
	where $g: \R^d \rightarrow \{0,1\}$ varies in the family of binary classifiers $\mathcal{G}$.
\end{Theorem}
$S$ is not $\varepsilon-$predictable from $X$ if $$ d_{TV}\left(\mathcal{L}\left(X|S=0\right), \mathcal{L}\left(X|S=1\right)\right) < 1-2\varepsilon$$
where $d_{TV}$ is the Total Variation distance. \vskip .1in
Hence fairness for all classifier $f$ is equivalent to the fact that 
$$ \min_{g \in \mathcal{F}}BER(g,X,S) = \frac{1}{2} $$ which is equivalent to $$ d_{TV}(\mu_0,\mu_1)=0,$$
where we have set $\mu_S=\mathcal{L}(X|S)$ for $S \in \{0,1\}$. Hence, perfect fairness for all classifiers in classification is equivalent to the fact that the distance between conditional distributions of the characteristics of individuals for the class defined by the different values of $S$ is null.

Consider transformations that map the conditional distributions to a joint distribution. Consider  $X \in  \mathbb{R}^d$ and $S \in \{0,1\}$. Let $T_S: \mathbb{R}^d\rightarrow \mathbb{R}^d, \ d\geqslant 1$ be a random transformation of $X$ such that $\mathcal{L}(T_0(X) \mid S=0)=\mathcal{L}(T_1(X) \mid S=1)$, and consider the transformed version $\tilde{X}=T_S(X)$. This transformation defines a way to {\it repair} the data in order to achieve fairness for all possible classifiers applied to these  repaired data $\tilde{X}=T_S(X)$. This maps transforms the distributions $\mu_S$ into their image by $T_S$, namely for all $S\in \{0,1\}$, ${\mu_S}_\sharp T_S:=\mu_S \circ T_S^{-1}$. Note that the choice of the transformation is equivalent to the choice of the target distribution $\nu_S= {\mu_S}_\sharp T_S$. Fairness is then achieved when the distance in Total Variations is equal to zero, which amounts to say that $T_0$ and $T_1$ maps the conditional distributions towards thew same distributions, hence $\nu_0=\nu_1$. \\
In this framework the price of fairness can be quantified as follows. For a given deformation $T_S$, set 
$$\mathcal{E}(T_S):=  \inf_{g \in \mathcal{G}} P(g(\tilde{X})\neq Y) - R_B(X,S).$$ The following theorem provides an upper bound for this price for fairness.
\begin{Theorem}(\cite{gordaliza2019obtaining})
	For each $s \in \{0,1\}$, assume that the function $\eta_s(x)=\p(Y=1 \mid X=x, S=s)$ is Lipschitz with constant $K_s >0$.
	Then, if $K=\max\{K_0,K_1\}$,
	\begin{equation*}
	\mathcal{E}(T_S) \leq 2 \sqrt{2}K \left(\sum_{s=0,1}\pi_sW_2^2(\mu_s, {\mu_s}_\sharp T_s)\right)^\frac{1}{2}.
	\end{equation*}
\end{Theorem}
Hence the minimal excess risk in this setting is achieved by minimizing previous quantity over possible transformations $T_S$. We thus obtain the following upper bound. 
\begin{align*}\label{equ:boundriskbarycenter}
\inf_{T_S} \mathcal{E}(T_S)  & \leq 2 \sqrt{2}K \inf_{T_S} \left(\sum_{s=0,1}\pi_sW_2^2(\mu_s, {\mu_s}_\sharp T_s)\right)^\frac{1}{2}\\
& \leq 2 \sqrt{2}K \inf_{\nu} \left(\sum_{s=0,1}\pi_sW_2^2(\mu_s, \nu)\right)^\frac{1}{2}\\
& =\sqrt{2}K \left(\sum_{s=0,1}\pi_sW_2^2(\mu_s, {\mu_B})\right)^\frac{1}{2}  
\end{align*}
where $\mu_B$ denotes the Wasserstein barycenter between $\mu_S$ with weight $\pi_S$ for $S\in \{0,1\}$. \\
Note that previous theorem can easily be extended to the case where $S$ takes multiple discrete values $S \in \{1,\dots,k\}$. In the case where $S$ is continuous, the same result holds using the extension of Wasserstein barycenter in \cite{le2017existence} and provided that conditional distributions $\mu_S$ are absolutely continuous with respect to Lebesgue measure.

\subsection{Price for fairness as Equality of Odds}
We study now the price for fairness meant as \textit{equality of odds}, which looks at the independence between the protected attribute and the outcome conditionally given the true value of the target, that is, the error of the algorithm.

\subsubsection{Regression}\label{subsubsec:PFEOregression}

Consider the regression framework detailed in section \ref{subsubsec:PFSPregression} and let $(X_1, S_1, Y_1), \ldots, (X_n, S_n, Y_n)$ be a sample of i.i.d. random vectors observed from $(X,S,Y)$. Denote by $\mathbb{X} \in \mathbb{R}^{n \times p}$ and $\mathbb{S} \in \mathbb{R}^{n\times 1}$ the matrices containing the observations of the non-sensitive and sensitive, respectively, features $X$ and $S$. We will assume standard normal independent errors $\varepsilon_1, \ldots, \varepsilon_n \sim \mathcal{N}(0,1)$. Then, we consider the linear normal model
\begin{equation}\label{equ:linearnormalmodel}
Y=f_{\beta_0,\beta}(X,S) + \varepsilon,
\end{equation}
where the errors are such that $\mathbb{E}(\varepsilon \mid (X,S))=0$, and the predictor 
\begin{equation}\label{eq:linearpred}
f_{\beta_0,\beta}(X,S)=\beta_0S+ \beta^TX,\ \beta_0 \in \mathbb{R},\ \beta \in \mathbb{R}^{p\times 1}
\end{equation}
is a linear combination of the sensitive and non-sensitive attributes. Then, the joint distribution of $(X,S,Y)$ is $(p+2)-$dimensional normal and we denote the vectors of means and the covariance matrices as follows
\begin{equation*}
(X,S,Y) \sim \mathcal{N} \left(\left[\begin{array}{c}
\mu_X\\
\mu_S\\
\mu_Y
\end{array}\right], \left[\begin{array}{ccc}
\Sigma_X & \Sigma_{XS} & \Sigma_{XY}\\
\Sigma_{XS}^T & \Sigma_S & \Sigma_{SY} \\
\Sigma_{XY}^T & \Sigma_{SY}^T & \Sigma_Y
\end{array}\right] \right)
\end{equation*}

We note that the \textit{equality of odds} criterion requires the linear fair predictor being independent of $S$ conditionally given $Y$, that is $$f_{\beta_0,\beta}(X,S) \independent S\mid Y,$$
which under the normal model is equivalent to the second order moment constraint
\begin{equation}\label{eq:condEOcov}
Cov(f(X,S), S \mid Y)=0.
\end{equation}
Hence, seeking for a fair linear predictor amounts to obtaining conditions on the coefficients $\beta_0,\beta$ for \eqref{eq:condEOcov} to hold. Since linear prediction can be seen as the most suitable framework for Gaussian processes, the relaxation of \eqref{eq:condEOcov} could be justified as being the appropriate notion of fairness when we restrict ourselves to linear predictors. Furthermote, linear predictors, especially under kernel transformations, are used in a wide array of applications. They thus form a practically relevant family of predictors where one would like to achieve non-discrimination. Therefore, in this section, we focus on obtaining non-discriminating linear predictors.

Now if we denote by $C_{S,X,Y} \in \mathbb{R}^{p\times 1}$ the \textit{vector of correction for fairness}
\begin{equation}
C_{S,X,Y}:=\left(\frac{\Sigma_{XS}\Sigma_Y-\Sigma_{SY}\Sigma_{XY}}{\Sigma_S\Sigma_Y-\Sigma_{SY}^2}\right),
\end{equation}
then the optimal fair \textit{equality of odds} predictor under the normal model can be exactly computed as in the following result, whose proof is set out in the Appendix \ref{app:optfairEOpred}.

\begin{Proposition}\label{propo:optfairEOpred}
	Under the normal model \eqref{equ:linearnormalmodel}, the optimal fair (\textit{equality of odds}) linear predictor of the form \eqref{eq:linearpred} is given as the solution to the following optimization problem
	\begin{align*}
	&\left(\hat{\beta}_{0,fair},\hat{\beta}_{fair}\right):=\argmin_{(\beta_0,\beta)\in \mathcal{F}_{EO}} \mathbb{E}\left[\left(Y-f_{\beta_0,\beta}(X,S)\right)^2\right]\\
	\mathcal{F}_{EO}&=\{(\beta_0,\beta)\in \R\times\R^p \ \text{such that} \ \beta^T\left(\Sigma_{XS}\Sigma_Y-\Sigma_{SY}\Sigma_{XY}\right)+\beta_0\left(\Sigma_S\Sigma_Y-\Sigma_{SY}^2\right)=0\} \nonumber.
	\end{align*}
	If moreover $Y$ and $S$ are not linearly dependent, it can be exactly computed as
	\begin{align*}
	&\hat{\beta}_{0,fair}= \hat{\beta}^T_{fair}C_{S,X,Y}\\
	&\hat{\beta}_{fair}=\Sigma_Z^{-1}\Sigma_{ZY},\\
	\text{where}& \nonumber\\\
	&\Sigma_Z=\Sigma_X+\Sigma_SC_{S,X,Y}C_{S,X,Y}^T+C_{S,X,Y}\Sigma_{XS}^T+\Sigma_{XS}C_{S,X,Y}^T \nonumber\\
	&\Sigma_{ZY}=\Sigma_{XY}+\Sigma_{SY}C_{S,X,Y} \nonumber.
	\end{align*}
\end{Proposition}

Note that the case where $Y$ and $S$ are linearly dependent corresponds to a totally unfair scenario that is not worth studying.\vskip .1in
We observe that, while condition \eqref{eq:condEOcov} is equivalent to \textit{equality of odds} in the normal setting, it is generally a weaker constraint. However, the problem of achieving perfect fairness as \textit{equalized odds} in a wider setup conveys computational challenges as discussed in \cite{woodworth2017learning}. They showed that even in the restricted case of learning linear predictors, assuming a convex loss function,
and demanding that only the sign of the predictor needs to be non-discriminatory, the problem of matching FPR and FNR requires exponential time to solve in the worst case. Motivated by this hardness result (see Theorem 3 in \cite{woodworth2017learning}), they also proposed a relaxation of the criterion of \textit{equalized odds} by a more tractable notion of non-discrimination based on second order moments. In particular, they proposed the notion of \textit{equalized correlations}, which indeed is generally a weaker condition than \eqref{eq:condEOcov}, but when considering the squared loss and when $(X,S,Y)$ are jointly Gaussian, it is in fact equivalent (and, subsequently, equivalent to \textit{equality of odds}). They also point out that for many distributions and hypothesis classes, there may not exist a non-constant, deterministic, perfectly fair predictor. Hence, we restrict ourselves here to the normal framework in which the computation of the optimal fair predictor is still feasible.


It is of interest to quantify the loss when imposing the fairness \textit{equality of odds} condition $(\beta_0,\beta)\in \mathcal{F}_{EO}$. This will be done comparing with the general loss associated to the minimizer
\begin{align}\label{optimal_NR}
\left[\hat{\beta}_{0}, \hat{\beta}^T\right]^T:=\argmin_{(\beta_0,\beta)\in \R \times \R^p} \mathbb{E}\left[\left(Y-f_{\beta_0,\beta}(X,S)\right)^2\right].
\end{align}

\begin{figure}[h]
		\begin{subfloatrow}
				\centering
			{\scriptsize
				\hspace{1cm}
		\ffigbox{
			\centering
			\begin{overpic}[scale=.4,unit=1mm]{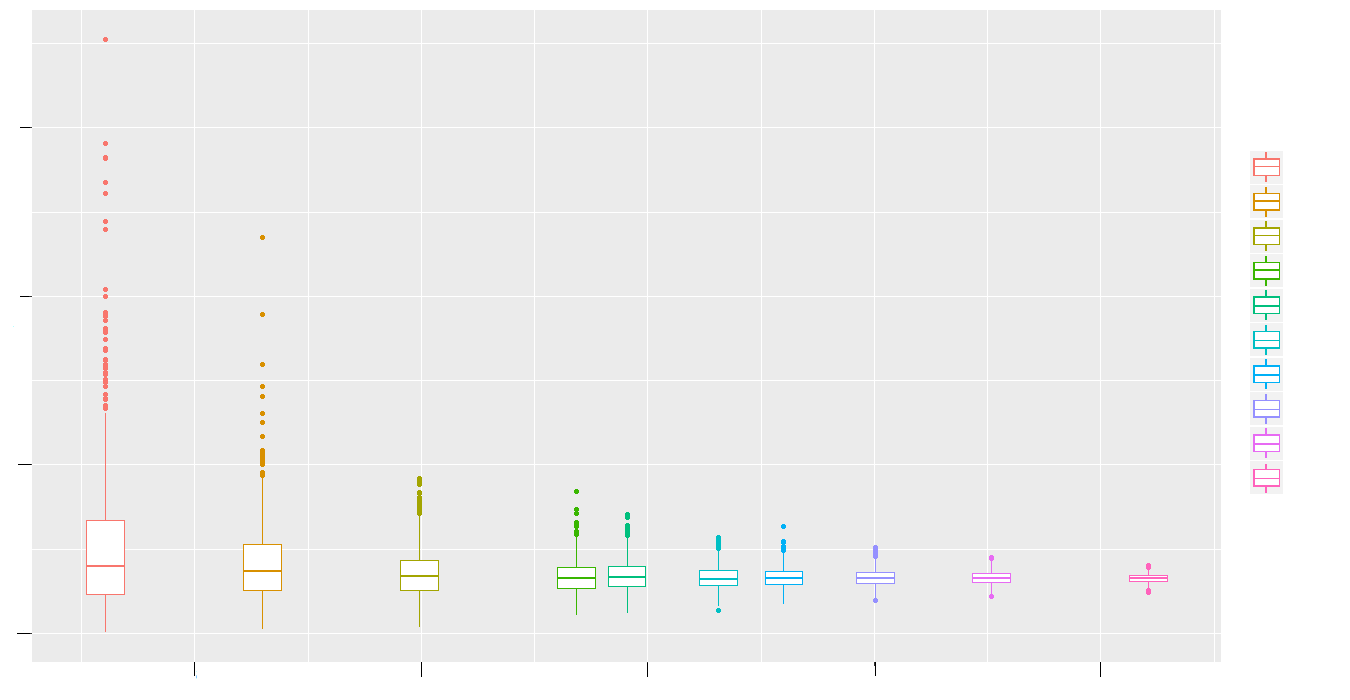}
				\put (-6,25){$\mathcal{E}_{\rm Fair}$}
				\put (-1,41.5){$6$}
				\put (-1,29){$4$}
				\put (-1,16.5){$2$}
				\put (-1,4){$0$}
				\put (42,-5){$\log$(size)}
				\put (14,-1){$5$}
				\put (31,-1){$6$}
				\put (47.5,-1){$7$}
				\put (64.5,-1){$8$}
				\put (81.5,-1){$9$}
				\put (95,42){size}
				\put (96,38.5){$100$}
				\put (96,36.25){$200$}
				\put (96,33.5){$400$}
				\put (96,31){$800$}
				\put (96,28.25){$1000$}
				\put (96,25.75){$1500$}
				\put (96,23.5){$2000$}
				\put (96,20.5){$3000$}
				\put (96,18){$5000$}
				\put (96,15.5){$10000$}
			\end{overpic}
		}{\label{fig:boxplot_averagemer}
		\vspace{0.5cm}
	\caption{Boxplot of the minimal excess risk computations}
}
	}
	\end{subfloatrow}
\vspace{0.5cm}
							{\scriptsize
	\ffigbox{
		\begin{subfloatrow}
			\ffigbox{
				\begin{overpic}[scale=.7,unit=1mm]{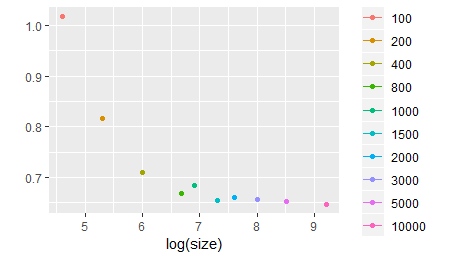}
					\put (-6,30){$\bar{\mathcal{E}}_{\rm Fair}$}
					\put (85,58){size}
					\end{overpic}
			}{\caption{average minimal excess risk}
				\label{fig:sim_mer_loss}
			}
			\ffigbox{
				\begin{overpic}[scale=.7,unit=1mm]{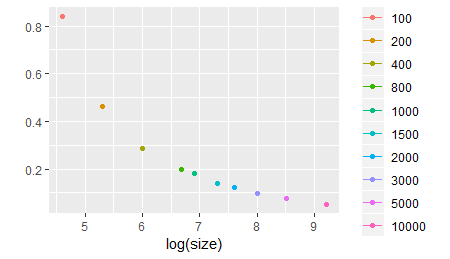}
						\put (-1,30){sd}
							\put (85,58){size}
			\end{overpic}
			}{\caption{standard deviation}
				\label{fig:sim_mer_err}
			}
		\end{subfloatrow}
	}{
		\caption{Minimal excess risk with $Cov(X_1,S)=0.1, \ Cov(X_2,S)=0.1$}
		\label{fig:sim_mer}
	}
}
\end{figure}
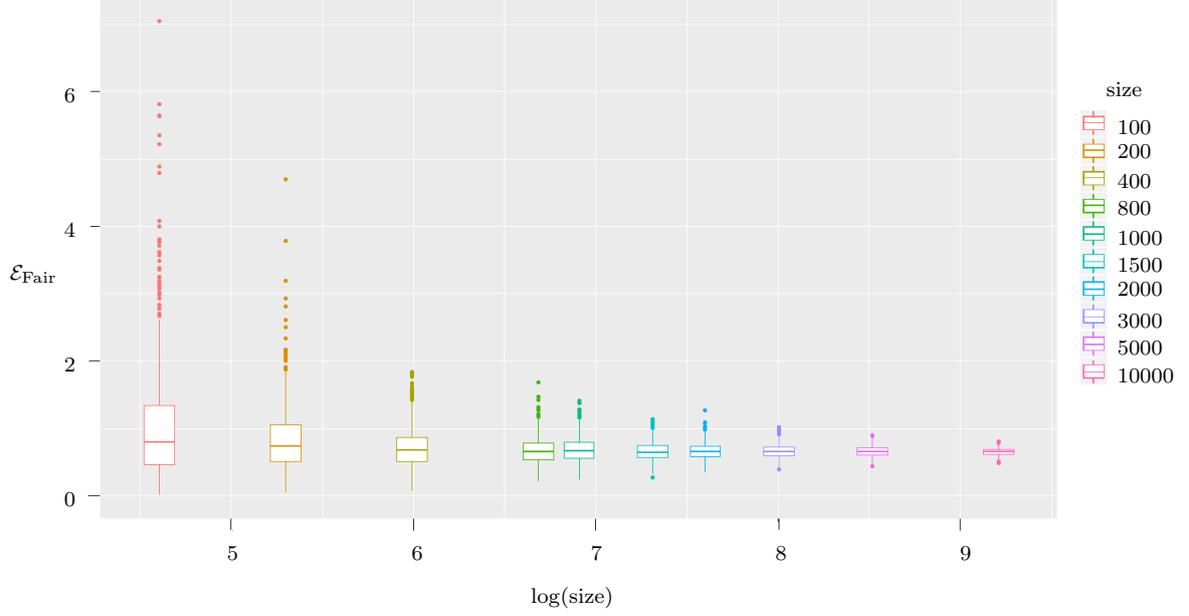
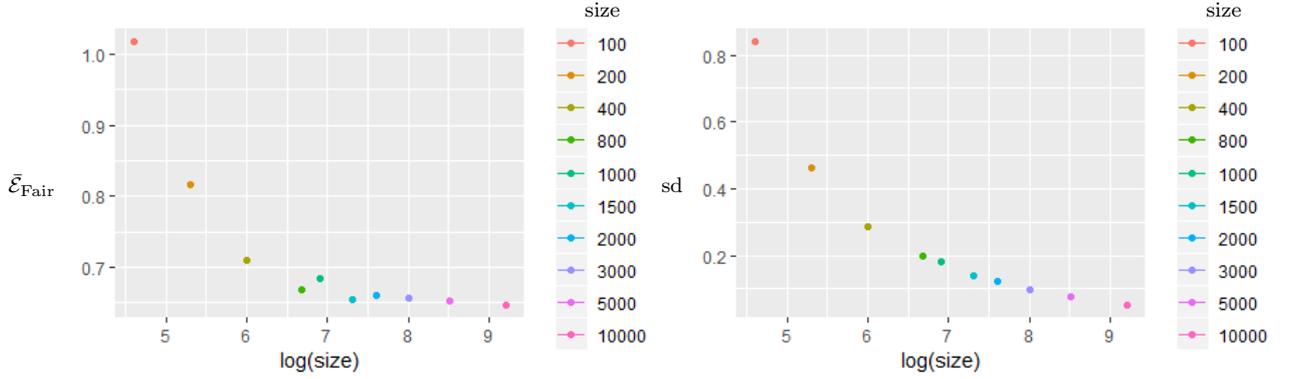

We have performed some simulations to obtain estimations of the minimal excess risk in \eqref{eq:theprice} when imposing \textit{equality of odds} under this gaussian linear regression framework. Precisely, we have considered $S \sim \mathcal{N}(0,10)$ and $X\in \mathbb{R}^2$, such that
\begin{equation*}
X \sim \mathcal{N} \left(\left[\begin{array}{c}
0\\
0
\end{array}\right], \left[\begin{array}{cc}
2 & 0\\
0& 3
\end{array}\right] \right).
\end{equation*}
The results of $1000$ replications of the experiment are shown in Figure \ref{fig:sim_mer}. There we present: (a) the average minimal excess risk; and its (b) standard deviation, as the sample size increases, taking particularly the values $(100,200,400,800,1000,1500,2000,3000,5000,10000)$. We observe that the estimation seems to converge.

\subsubsection{Classification}\label{subsubsec:PFEOclass}
We consider again the classification setting where we wish to predict a binary output label $Y \in \{0,1\}$ from the pair $(X,S)$. In this section, we obtain the fair optimal classifier in the sense of \textit{equality of odds} in the particular case where $S$ is also binary. We assume moreover that both the marginals and the joint distribution of $(S,Y)$ are non-degenerate, that is $\mathbb{P}(Y=1)\in (0,1), \ \mathbb{P}(S=1)\in (0,1)$ and $\mathbb{P}(Y=1,S=1) \in (0,1)$. There are some other works dealing with the computation of Bayes-optimal classifiers under different notions of fairness. In \cite{menon2018cost} \textit{statistical parity} and \textit{equality of oportunity} are the considered constraints. Our approach here extends the proposed in \cite{chzhen2019leveraging}, where fairness is defined by the weaker notion of \textit{equality of opportunity} that requires just the equality of true posisitive rates across both groups.

An optimal fair classifier is formally defined here as the solution to the risk minimination problem over the class $\mathcal{F}_{EO}$ of binary classfiers satisfying the \textit{equality of odds} conditions, that is
\begin{align*}
&g^* \in \argmin_{g \in \mathcal{F}_{EO}}\mathcal{R}(g), \ \text{where}\\
&\mathcal{F}_{EO}:=\{g \in \mathcal{G}: \mathbb{P}(g(X,S)=i \mid Y=i, S=0)=\mathbb{P}(g(X,S)=i \mid Y=i, S=1), i=0,1\}.
\end{align*}

In order to establish the form of such minimizer, we introduce the following assumption on the regression function.

\begin{Assumption}\label{ass:optEOclass} For each $s\in \{0,1\}$ we require the mapping $t\in\mathbb{P}(\eta(X,S)\leq t \mid S=s)$ to be continuous on $(0,1)$, where for all $(x,s) \in \R^d\times\{0,1\},$ we let the regression function
	\begin{equation}\label{def:regressionfunction}
	\eta(x,s):=\mathbb{P}(Y=1 \mid X=x, S=s)=\mathbb{E}\left[Y\mid X=x, S=s\right].
	\end{equation}
\end{Assumption}
The following result establishes that the optimal \textit{equalized odds} classifier is obtained recalibrating the Bayes classifier $g_B(X,S)=\mathbbm{1}_{\{\eta(X,S)\geq 1/2\}}$, and its proof is included in the Appendix \ref{app:optfairEOclass}.
\begin{Proposition}[Optimal Rule]\label{propo:optEOclass}
	Under Assumption \ref{ass:optEOclass}, an optimal classifier $g^*$ can be obtained for all $(x,s) \in \R^d\times\{0,1\}$ as
	\begin{align*}
	g^*(x,1)&=\mathbbm{1}_{\{1\leq 2\eta(X,1)-\theta_1^*\frac{\eta(X,1)}{\mathbb{P}(Y=1, S=1)}+\theta_0^*\frac{1-\eta(X,1)}{\mathbb{P}(Y=0, S=1)}\}} \\
	g^*(x,0)&=\mathbbm{1}_{\{1\leq 2\eta(X,0)+\theta_1^*\frac{\eta(X,0)}{\mathbb{P}(Y=1, S=0)}-\theta_0^*\frac{1-\eta(X,0)}{\mathbb{P}(Y=0, S=0)}\}},
	\end{align*}
	where $(\theta_0^*, \theta_1^*) \in \R^2$ is determined from equations
	\begin{align*}
	\frac{\mathbb{E}_{X \mid S=1}\Big[\eta(X,1)g^*(X,1)\Big]}{\mathbb{P}(Y=1\mid S=1)}&=\frac{\mathbb{E}_{X \mid S=0}\Big[\eta(X,0)g^*(X,0)\Big]}{\mathbb{P}(Y=1\mid S=0)} \\
	\frac{\mathbb{E}_{X \mid S=1}\Big[(1-\eta(X,1))g^*(X,1)\Big]}{\mathbb{P}(Y=0\mid S=1)}&=\frac{\mathbb{E}_{X \mid S=0}\Big[(1-\eta(X,0))g^*(X,0)\Big]}{\mathbb{P}(Y=0\mid S=0)}.
	\end{align*}
\end{Proposition}
\begin{Remark}
	Note that if $\theta_0^*=0$ we recover the optimal fair \textit{equality of opportunity} classifier in \cite{chzhen2019leveraging}. If moreover $\theta_0^*=1$ the above defined is the classical Bayes rule.
\end{Remark}

We have quantified the cost with respect to the loss of the generalization error needed to ensure fairness in machine learning for classification and regression. If this price appears too high for the practitioner, the notion of fairness has to be weakened into a quantitative measure that can be adjusted for a trade-off between accuracy to the observations and fairness.

\section{Quantifying fairness in machine learning}

The importance of ensuring fairness in algorithmic outcomes has raised the need for designing procedures to remove the potential presence of bias. Yet building perfect fair models may lead to poor accuracy: changing the world into a fair one with positive action might decrease the efficiency defined as its similarity to the uses monitored through the test sample. While in some fields of application, it is desirable to ensure the highest possible level of fairness; in others, including Health Care or Criminal Justice, performance should not be decreased since the decisions would have serious implications for individuals and society. Hence, when perfect fairness requires to pay a too great price, resulting in poor generalization errors with respect to the unfair case, it is natural not to impose this strict condition but rather weaken the fairness constraint. In other words, it is of great interest to set a trade-off between fairness and accuracy, resulting in a relaxation of the notion of fairness that is frequently presented in the literature as \textit{almost} or \textit{approximate fairness}. To this aim, most methods approximate fairness desiderata through requirements on the lower order moments or other functions of distributions corresponding to different sensitive attributes.

From a procedural viewpoint, methods for imposing fairness are roughly divided in the literature into three families. Methods in the first family consist in pre-processing the data or in extracting representations that do not contain undesired biases (see e.g. \cite{beutel2017data, calders2009building, calmon2017optimized, chierichetti2017fair, edwards2015censoring, feldman2015computational, FFMSV, fish2015fair, gordaliza2019obtaining, johndrow2019algorithm, kamiran2009classifying, kamiran2010classification, kamiran2012data, madras2018learning, song2018learning, zemel2013learning}), which can then be used as input to a standard machine learning model. Methods in the second family, also referred to as in-processing, aim at enforcing a model to produce fair outputs through imposing fairness constraints into the learning mechanism. Some methods transform the constrained optimization problem via the method of Lagrange multipliers (see e.g. \cite{agarwal2018reductions, berk2017convex, corbett2017algorithmic, cotter2018training, kearns2018preventing, narasimhan2018learning, ZVGG, zafar2019fairness}) or add penalties to the objective (see e.g. \cite{bechavod2017penalizing, donini2018empirical, dwork2018decoupled, fukuchi2015prediction, hebert2018calibration, kamiran2012decision, kamishima2012fairness, kilbertus2017avoiding, komiyama2018nonconvex, madras2018predict, mary2019fairness, nabi2018fair, narasimhan2018learning, oneto2019taking, speicher2018unified, yona2018probably}), others use adversary techniques to maximize the system ability to predict the target while minimizing the ability to predict the sensitive attribute \cite{zhang2018mitigating}. Methods in the third family consist in post-processing the outputs of a model in order to make them fair (see e.g. \cite{adler2018auditing, ali2019loss, chzhen2019leveraging, doherty2012information, feldman2015computational, fish2016confidence, hajian2012injecting, hardt2016equality, kim2019multiaccuracy, kusner2017counterfactual, noriega2019active, pedreschi2009measuring}).

As noticed in \cite{oneto2020recent}, this grouping is imprecise and non exhaustive. In the following we describe more deeply two different families of methods, which are non-mutually exclusive. First a group of in-processing methods which can be seen as a fair risk minimization problem and includes the majority of the contributions.  On the other hand, a second category of methods based on optimal transport, which correspond mostly to pre or post processing approaches, since it is the preferred tool in this thesis for fair learning.


\subsection{Fairness through Empirical Risk Minimization}
We recall that the aim of a supervised machine learning algorithm is to learn the relationships between input characteristic variables and a target variable in order to forecast new observations. In the fair learning setting, we observe $(X_1,S_1,Y_1),\dots, (X_n,S_n,Y_n) $ i.i.d observations drawn from an unknown distribution $\mathbb{P}$. Set the empirical distribution $\mathbb{P}_n=\frac{1}{n} \sum_{i=1}^n \delta_{X_i,S_i,Y_i}$. An almost-fair model will be obtained by minimizing the empirical risk $$R_n(f)=\frac{1}{n} \sum_{i=1}^n\ell (Y_i,f(X_i,S_i)),$$ with $\ell  :  \: (Y,\hat{Y}) \mapsto \ell(Y,\hat{Y}) \in \mathbb{R}^+$ a certain loss function measuring the quality of the prediction, and where the influence of the protected variable $S$ in the forecast $\hat{Y}$ should be controlled. We note that such influence must be nule in the case of perfect fairness and could be imposed by minimizing over a class $\mathcal{F}_{fair}$ satisfying certain stringent conditions. The classes $\mathcal{F}_{SP}$ or $\mathcal{F}_{EO}$, defined respectively in \eqref{defi:FairClassSP} and \eqref{defi:FairClassEO}, are two possibilities for the minimization. In general, a relaxation of the problem would enable control on the level of fairness of the learnt algorithm. This is proposed in the majority of the papers either by
\begin{itemize}
	\item[(i)] thresholding full-type fairness conditions, that is
	\begin{equation}\label{app:fairconstraints}
	\min_{f\in \mathcal{F}}R_n(f)  \quad \text{such that} \ \delta(f(X,S),S,Y)\leq \varepsilon,
	\end{equation}
	where $\delta$ is a measure of dependency (with $\delta(f(X,S),S,Y)=0$ in the perfect-fair case) and $\varepsilon >0$ represents the level of fairness; or
	\item[(ii)] directly introducing the independency as a penalty into the objective
	\begin{equation}\label{app:fairpenalty}	
	\min_{f\in \mathcal{F}} \left\{ R_n(f)  +  \lambda \delta(f(X,S),S,Y) \right\},
	\end{equation}
	where $\lambda>0$ balances the contribution of both terms to get a trade-off between the bias and the efficiency of the algorithm.
\end{itemize}

Yet the main question becomes in how to select the notion of independency measured above through the function $\delta$. Several choices exist in the literature. According to the division of perfect fairness notions proposed in section \ref{subs:PFairness}, \textit{almost fairness} requires quantifying the dependence between the distribution of the protected variable $S$ and 
\begin{itemize}
	\item[(i)] either the distribution of the forecast $\widehat{Y}$, or the conditional distribution of the forecast given the true value $\widehat{Y} | Y $,
	\item[(ii)] or the expectation $\mathbb{E} \Phi(\widehat{Y})$ or  $\mathbb{E} (\Phi(\widehat{Y}) |Y)$, through a chosen function $\Phi: \mathbb{R} \rightarrow \mathbb{R}$. If $\Phi(X,Y,S)=\ell(f(X),Y)$ Sobol index of $\Phi$ is a good criterion :  sensivity analyis is a fairness criterion.
\end{itemize}

Both points of view correspond to choices that can be  made. In the following, we review how this framework summarizes most of the recent papers dealing with almost fairness.

\subsubsection{Imposing conditions on the distributions}\label{subsubsec:conddist}

The first set of approaches to get fair predictive behaviour by adding constraints through conditions over the distributions has been studied in several papers. Depending on the basis of such conditions, the main proposals can be organised as follows:
\begin{enumerate}
	\item[(a)] \textbf{Distance-based constraints}. According to the definition of fairness as independency criterion, this category of approaches aims at quantifying the distance between the probability distributions: 
	\begin{enumerate}
		\item[(i)] $\mathcal{L}(\widehat{Y} | S=s),$ for all $s \in \mathcal{S} $; or $\mathcal{L}(\widehat{Y} \times S)$ and $\mathcal{L}(\widehat{Y} ) \times \mathcal{L} (S)$, if \textit{statistical parity} is considered.
		\item[(ii)] $\mathcal{L}(\widehat{Y} | Y,S=s),$ for all  $s \in \mathcal{S}$, regarding to \textit{equality of odds}.
	\end{enumerate}
	The majority of the papers in this line of work considered Wasserstein distances and we summarize the main contributions hereafter. In \cite{jiang2019wasserstein} two different approaches to achieve \textit{statistical parity} with Wasserstein-1 distance are proposed. First, a fast and practical approximation methodology to post-process the model outputs by enforcing the density functions of probabilities $\mathcal{L}(\hat{Y}\mid S=s)$ corresponding to groups of individuals with different sensitive attributes to coincide with their Wasserstein-1 barycenter distribution. Then, a penalization approach to binary logistic regression that aims at finding the model parameters minimizing the logistic loss under the constraint of small Wasserstein-1 distances between the empirical counterparts of measures $\mathcal{L}(\hat{Y}\mid S=s)$ and their empirical barycenter. 
	
	Wasserstein-type constraints for building fair classifiers has also been considered in \cite{serrurier2019fairness}. They provided algorithms which can incorporate both notions of fairness through 1-Wasserstein distance-based contraints. Yet sharing some similarities with \cite{edwards2016towards}, their approach is more flexible and enables to solve wider classes of fairness problems based on different adversarial architecture resulting in more suited loss functions. Neural networks are used to manage a large variety of input data structure (e.g. images) as well as output labels (multiclass, regression, images...). Their Wasserstein approximation using fairness benchmark datasets outperformed both classical fair algorithms (e.g fair SVM) as well as similar adversarial architectures based on Jensen or GAN losses (see references in the paper for more details.)
	
	In \cite{risser2019using} algorithmic fairness is promoted by imposing closeness with respect to quadratic Wasserstein distance between the scores used to build an automatic decision rule. This regularization constraint is built with a deep neural network.
	
	Specifically the concept of the barycenter in optimal transport theory is used in the recent paper \cite{zehlike2020matching} to maximize decision maker utility under the chosen fairness constraints. They proposed the \textit{Continuous Fairness Algorithm} which enables a continuous interpolation between different fairness definitions. This algorithm is able to handle cases of multi-dimensional discrimination of	certain groups on grounds of several criteria. They included examples of credit applications, college admissions and insurance contracts; and mapped out the legal and policy implications of their approach.

	\item[(b)] \textbf{Information theory-based contraints}. 	
	First contributions to this approach in the context of fair supervised learning started with the work of \cite{kamishima2011fairness}, who designed an unfairness penalty term based on \textit{statistical parity} criterion (referred to in their paper as \textit{indirect prejudice}), which restricts the amount of mutual information between the prediction and the sensitive attribute.
	More precisely, they add a fairness regularization term in the objective function that penalizes the mutual information between the sensitive feature and the classifier decisions. In this way, this method treats the mutual information as the unfairness proxy. Their technique is only limited to the logistic regression classification model. Later in \cite{kamishima2012fairness} they used normalised MI to asses fairness in their \textit{normalised prejudice index (NPI)}. Their focus is on binary classification with binary sensitive attributes, and the NPI is based on the independence fairness criterion. In such setting, mutual information is readily computable empirically from confusion matrices. This work is generalised in \cite{fukuchi2015prediction} for use in regression models by using a neutrality measure, which is shown to be equivalent to the independence criterion. They then use this neutrality measure to create inprocessing techniques for linear and logistic regression algorithms. Similarly, \cite{ghassami2018fairness} take an information theoretic approach to creating an optimisation algorithm that returns a predictor score that is fair with respect to the \textit{equalized odds} criterion. 
	
	An information theory motivated framework is also proposed in \cite{song2018learning} where the goal is to maximize what they called the \textit{expressiveness} of representations of the data while satisfying certain fairness constraint. Expressiveness, as well as \textit{statistical parity}, \textit{equalized odds} and \textit{equalized opportunity}, are expressed in terms of mutual information, and tractable upper and lower bounds of these mutual information objectives are obtained. A conexion between them and existing objectives such as maximum likelihood, adversarial training (Goodfellow et al., 2014), and variational autoencoders (Kingma and Welling, 2013; Rezende and Mohamed, 2015) is also presented. Their contribution serves as a unifying	framework for existing work \cite{zemel2013learning, edwards2016towards, madras2018learning} on learning fair representations, being the first to provide direct user control over the fairness of representations through fairness constraints that are interpretable by non-expert users. 
	
	In the regression setting, measuring group fairness criteria is computationally challenging, as it requires estimating	information-theoretic divergences between conditional probability density functions. Recently \cite{steinberg2020fast}  introduced fast approximations of the \textit{statistical parity}, \textit{equality of odds} and \textit{predictive parity} (there referred to as \textit{independence}, \textit{separation} and \textit{sufficiency}, respectively; following ...) fairness criteria for regression models from their (conditional) mutual information definitions, and used such approximations as regularisers to enforce fairness within a regularised risk minimisation framework. 
	
	
	\item[(c)] \textbf{Kernel theory-based constraints}.
	Regularization is one of the key concepts in modern supervised learning, which allows imposing structural assumptions and inductive biases onto the problem at hand. It ranges from classical notions of sparsity, shrinkage, and model complexity to the more intricate regularization terms which allow building specific assumptions about the predictors into the objective functions, such as smoothness on manifolds \cite{belkin2006manifold}. Such regularization
	viewpoint for algorithmic fairness was presented in \cite{kamishima2012fairness} in the context of classification, and was extended to regression and unsupervised dimensionality reduction problems with kernel methods in \cite{perez2017fair}. The latter falls within the framework
	of \textit{statistical parity} and was the first work that considered this notion with continuous labels. They proposed kernel machines to exploit cross-covariance operators in Hilbert spaces. In particular, independence between predictor and sensitive variables is imposed by employing a kernel dependence measure, namely the Hilbert-Schmidt Independence Criterion (HSIC) \cite{gretton2005kernel}, as a regularizer in the objective function.
	
	Extentions of this work are presented in \cite{li2019kernel} where a general framework of empirical risk minimization with fairness regularizers and their interpretation is given. Secondly, they derived a Gaussian Process (GP)
	formulation of the fairness regularization framework, which allows uncertainty
	quantification and principled hyperparameter selection. Finally, we introduce
	a normalized version of the fairness regularizer which makes it less sensitive to
	the choice of kernel parameters. We demonstrate how the developed fairness
	regularization framework trades off model’s predictive accuracy (with respect
	to potentially biased data) for independence to the sensitive covariates. It is
	worth noting that, in our setting, a function which produced the labels is not
	necessarily the function we wish to learn, so that the predictive accuracy is
	not necessarily a gold-standard criterion.
\end{enumerate}

\subsubsection{Imposing conditions on the expectation}
On the other side, reinforcement of fair algorithmic behaviour has been also proposed by requiring conditions on the expected forecast in a large number of papers. More precisely, depending, on the one hand, on the desirable metric of fairness (as discussed in section \ref{s:definition}); and on the other, on the nature of the target $Y$ and the protected attribute $S$, the dependence measure $\delta$ is set out to control different kinds of indexes. We note that this control could be imposed following either \eqref{app:fairconstraints} or \eqref{app:fairpenalty}.
\begin{enumerate}
	\item For \textit{statistical parity}, if $Y,S \in \{0,1\}$, conditions on the probabilities of success across groups $\mathbb{P}(f(X,S) =1 |S)$ are considered, being the \textit{mean difference score} 
	\begin{equation}\label{eq:MD}
	\mathbb{P}(f(X,S) =1 |S=1 )-\mathbb{P}(f(X,S) =1 |S =0),
	\end{equation}
	which was first introduced in \cite{calders2010three}, and the disparate impact \eqref{def:DIclassifier} the preferent choices in the literature. These are generalized to conditions on the expectation $\mathbb{E} ( f(X,S)|S)$ (or $\mathbb{E} ( \ell(f(X,S),Y)|S)$, with $\ell$ a loss function) or, from a sensitivity analysis point of view, on the variances ${\rm Var} (\mathbb{E} ( f(X,S)|S))$ (or ${\rm Var} (\mathbb{E} ( \ell(f(X,S),Y)|S))$).
	
	\item For \textit{equality of odds}, if $Y,S \in \{0,1\}$, then the goal is similar as before but taking into account the true values of the target $Y$. Namely, the differences between TPR and FPR, that is
	\begin{equation}\label{eq:TPNRD}
	\mathbb{P}(f(X,S) =i |Y=i,S=1 )-\mathbb{P}(f(X,S)=i |Y=i,S=0), \ \text{for} \ i=0,1,
	\end{equation}
	are usually considered. Besides, in a less demanding way, others focus on the difference between the overall accuracies
	\begin{equation}\label{eq:OAD}
	\mathbb{P}(Y \neq \hat{Y} \mid S=0)- \mathbb{P}(Y \neq \hat{Y} \mid S=1).
	\end{equation}
	In a wider setup, this approach is extended to conditions on the expectation $\mathbb{E} (f(X,S)|Y,S)$  (or $\mathbb{E} (\ell (f(X,S),Y)|Y,S)$) or the variance ${\rm Var} (\mathbb{E}(f(X,S)|Y,S)$ (or ${\rm Var} (\mathbb{E} (\ell (f(X,S),Y)|Y,S)$).
\end{enumerate}

Given this overview summarizing the majority of proposals for relaxing the notion of fairness through conditions on the input and output distributions of the algorithm, we cite some of the main contributions to this approach. One of the first was the work of \cite{zemel2013learning} which, based on \cite{dwork2012fairness}, combined pre-processing and inprocessing by jointly learning a ‘fair’ representation of the data and the classifier parameters. The joint representation is learnt using a multi-objective loss function that ensures that (i) the resulting representations do not lead to disparate impact, (ii) the reconstruction loss from the original data and intermediate representations is small and (iii) the class label can be predicted with high accuracy. This approach has two main limitations: i) it leads to a
non-convex optimization problem and does not guarantee optimality, and ii) the accuracy of the classifier depends on the dimension of the fair representation, which needs to be chosen rather arbitrarily. Inspired by \cite{zemel2013learning}, the methods of \cite{edwards2016towards} and \cite{madras2018learning} also aim at learning fair representations of the data.

In \cite{ZVGG} methods for training decision boundary-based classifiers without \textit{disparate mistreatment} (recall \eqref{def:DM}) are described, with further extensions to existing notions \textit{disparate treatment} and \textit{disparate impact} in \cite{zafar2017parity}. Their proposals, as well as the results of several experiments and applications to well-known real datasets, have been collected later in \cite{zafar2019fairness}. They noticed that taking in the above formulation \eqref{app:fairconstraints} the dependence measure $\delta$ in terms of the accuracies in \eqref{eq:OAD}, and similarly for \eqref{eq:TPNRD}, ensures that the classifier chooses the optimal decision boundary within the space of fair boundaries specified by the constraints but yields to a very challenging problem. The reason is two-fold: first, the fairness constraints lead to non-convex formulations; and second, the probabilities defining such constraints are function having saddle points, which further complicates the procedure for solving non-convex optimization problems \cite{dauphin2014identifying}. Therefore, they proposed a relaxation of these (non-convex) fairness constraints into proxy conditions, each in the form of a convex-concave (or, difference of
convex) function using a covariance measure of decision boundary fairness. They design fair logistic regression classifiers and linear and nonlinear SVMs as examples and heuristically solve the resulting optimization problem for a convex loss function.
Adding constraints to the classification model is also in the line of work of \cite{goh2016satisfying}, \cite{woodworth2017learning} and \cite{quadrianto2017recycling}. While the constraints are similar to those in \cite{zafar2019fairness}, the first two are only limited to a single specific loss function and the third one to a single notion of unfairness.

Another approach in pursuit of fairness as \textit{equality of odds} in binary classifiers learned over individuals from two populations is presented in \cite{bechavod2017penalizing}. They validate the ability of such approach to achieve both fairness and high accuracy, implementing and testing it on multiple datasets from the fields of criminal risk assessment, credit, lending, and college admissions. Later in \cite{agarwal2018reductions} both \textit{statistical parity} and \textit{equalized odds} conditions are viewed as a special case of a general set of linear constraints. Based on that, the minimization problem is shown to be reduced to a sequence of cost-sensitive classification problems, whose solutions yield a randomized classifier with the lowest (empirical) error subject to the desired constraints.

In \cite{menon2018cost} \textit{disparate impact} and \textit{mean difference} indexes are related to cost-sensitive risks and the tradeoffs between performance of  in the problem of learning with these fairness constraint are studied. They showed that the optimal classifier for these cost-sensitive measures is an instance-dependent thresholding of the classprobability function, and quantify the degradation in performance by a measure of alignment of the target and sensitive variable. They also use such analysis to derive a simple plugin approach for the fairness problem. Finally, in the classification setting we metion also \cite{kearns2018preventing}, who considered the problem of learning binary classifiers subject to \textit{equal opportunity} and \textit{statistical parity} constraints when the number of protected groups is large.

In the fair regression framework, \cite{ZVGG} suggested a relaxed notion of non-discrimination based on first order moments $$\mathbb{E}(\hat{Y}|Y=y, S=0)=\mathbb{E}(\hat{Y}|Y=y, S=1)$$ and proposed optimizing a convex loss subject to an approximation of this constraint. With a similar aim, in previously cited paper \cite{woodworth2017learning} (see section \ref{subsubsec:PFEOregression}) they proposed a relaxation of the criterion of \textit{equalized odds} by a more tractable notion of non-discrimination based on second order moments. In particular, they proposed the notion of \textit{equalized correlations}. Later, in \cite{agarwal2019fair} the fair regression problem is studied in a predictive setting where $\mathcal{X}$ could be continuous and high-dimesional, $\mathcal{S}$ is discrete, and $\mathcal{Y}\subseteq \left[0,1\right]$ could be discrete (but embedded in $\left[0,1\right]$) or continuous. Two different constraints in the minimization \eqref{app:fairconstraints} are considered in this work. Firstly, a relaxation of \textit{statistical parity} is proposed as, for all $z \in \left[0,1\right]$ and all $s \in \mathcal{S}$,
\begin{equation}
|\mathbb{P}(f(X) \geq z \mid S=s)-\mathbb{P}(f(X) \geq z|\leq \varepsilon_s,
\end{equation}
where the slack $\varepsilon_s >0$ bounds the allowed departure of the CDF of $f(X)$ conditional on $S=s$ from the CDF of $f(X)$. Note that the protected variable $S$ is not explicitely considered as input. The difference between CDFs is measured in the $\infty-$norm corresponding to the Kolmogorov-Smirnov statistic.  On the other hand, they also propose to guarantee fairness through the criteria \textit{bounded group loss}
\begin{equation}
\mathbb{E} ( \ell(f(X),Y)|S) \leq \varepsilon_s
\end{equation}
where, in fact, the threshold is uniform for all the classes in the definition, but, for the sake of flexibility, it is allowed to specify different bounds $\varepsilon_s >0$ for each attribute value in the loss minimization. Hence, fair regression with \textit{bounded group loss} minimizes the overall loss, while controlling the worst loss on any protected group. By Lagrangian duality, this is equivalent to minimizing the worst loss on any group while maintaining good overall loss (referred to as \textit{max-min fairness}). Unlike \textit{overall accuracy equality} in classification \cite{dieterich2016compas}, which requires the losses on all groups to be equal, they claimed that \textit{bounded group loss} does not force an artificial decrease in performance on every group just to match the hardest-to-predict group. They also generalized their approach to randomized predictors to achieve better fairness-accuracy trade-off.

We finally cite the recent algorithm in \cite{oneto2020recent} called \textit{General Fair Empirical Risk Minimization (G-FERM)} that generalizes the \textit{Fair Empirical Risk Minimization} approach introduced in \cite{donini2018empirical}. In this work, they also specify the method for the case in which the underlying
space of models is a RKHS and show how the in-processing G-FERM approach described above can be translated into a pre-processing approach.

\subsection{Fairness through Optimal Transport}
Most methods obtain fair models by imposing approximations of fairness desiderata
through constraints on lower order moments or other functions of distributions corresponding to different sensitive attributes (this is also what most popular fairness definitions require). As observed in \cite{oneto2020recent}, whilst facilitating model design, not imposing constraints on the	full shapes of relevant distributions can be problematic. One existing approach that does work this way propose to match distributions corresponding to different sensitive attributes either in the space of model outputs or in the space of model inputs (or latent representations of the inputs) using optimal transport theory, which correspond to post and pre-processing methods, respectively. We note that the in-processing methods based on optimal transport are those imposing constraints in terms of the Wasserstein distance and have already been described above (see in section \ref{subsubsec:conddist}(a)).

The idea of the pre-processing based methods to obtain fair treatment consists in blurring the value of the protected class by transporting the original distribution of the input, conditionally to this value, towards their Wasserstein's barycenter. It was first considered in the binaty classification problem in \cite{FFMSV}, \cite{johndrow2019algorithm} or \cite{DBLP:journals/corr/abs-1712-07924}, and later improved in \cite{gordaliza2019obtaining}. In this work, the choice of the weighted Wasserstein's barycenter with respect to the weights of the protected clased is formally justified (see Theorem 4.3.3.) in terms of the minimal excess risk when considering the classfier trained from the repaired data. Moreover, they propose to set an accuracy-fairness trade-off through a partial repair approach called \textit{random repair}, which it is shown to outperform the previous geometric repair in \cite{FFMSV}.

The work in \cite{chiappa2020general, jiang2019wasserstein} presents an approach to fair classification and regression that is applicable to many fairness criteria. In particular, they introduce the notion of \textit{Strong Demographic Parity}, which extends the \textit{statistical parity} to a fair multi-classification and regression problem. 
Based on that, in \cite{oneto2020recent} they derived a simple post-processing method withing this framework to achieve \textit{Strong Demographic Parity} by transporting distributions to their Wasserstein barycenter. They also propose a partial transportation for setting a fairness-accuracy trade-off called the \textit{Wasserstein 2-Geodesic} method.

\section{Conclusions}
In this paper, we have presented a review of mathematical models designed to handle the issue of bias in machine learning. Due to the large number of definitions, we have propose a probabilistic framework to understand the relationships between fairness and the notion of independence or conditional independence. Hence imposing fairness is here modeled as imposing independence with respect to the sensitive variable and constraints are naturally driven by the choice of different measures for this independence. Within this framework, it becomes thus possible to give another insight at several notions of fairness and also to quantity their effect on the decision rule. In particular, we can defined and then compute in some cases the so-called price for fairness to quantify the real impact of fairness constraint on the behavior of a machine learning algorithm.  This study provides a better understanding of fair learning, each different definition of fairness leading to different behaviors that can be compared in some cases. Yet many cases remain open to further research to obtain a full theoretical framework of fair learning.

Moreover, we point out that we did not consider in this study many new interesting points of view on fairness that deserve a specific study. In very particular, understanding fairness from a causal point of view or using counter-examples as in \cite{loftus2018causal} and \cite{kusner2017counterfactual} or \cite{black2020fliptest} could provide another interpretation for fairness in machine learning.
	
\section{Appendix}
\subsection{Proofs of section \ref{subsec:relation}}\label{app:relation}

\textbf{Proof of Proposition \ref{prop:SPvsEO}} Observe that if $S \independent \hat{Y}$ and $\hat{Y} \independent Y \mid S $ then either $S \independent Y$ or $ \hat{Y} \independent Y$.

$\hfill\Box$\\

\noindent\textbf{Proof of Proposition \ref{prop:SPvsPP}} It suffices to observe that if $S \not\independent Y$ and $S \independent Y \mid\hat{Y} $ then $S \not\independent \hat{Y}$.

$\hfill\Box$\\

\noindent\textbf{Proof of Proposition \ref{prop:PPvsEO}} $S \independent \hat{Y}\mid Y$ and $S \independent Y \mid\hat{Y} $ implies $S \independent (\hat{Y}, Y)$, and then $S \independent Y$.

$\hfill\Box$\\

\subsection{Proofs of section \ref{subsubsec:PFEOregression}}\label{app:optfairEOpred}

We start recalling some facts about Gaussian random variables.
\begin{Proposition}\label{prop:gaussian}
	If $(U,V,W)$ are jointly Gaussian, then
	\begin{itemize}
		\item Conditional expectation $\mathbb{E}(U|V)$ is linear in $V$ and is given by
		\begin{equation*}
		\mathbb{E}(U|V)=\mathbb{E}(U)+\Sigma_{U,V}\Sigma^{-1}_{V}(V-\mathbb{E}(V))
		\end{equation*}
		\item Conditional covariance $\Sigma_{(U,V)|W}$ does not depend on $W$ and is given by
		\begin{equation*}
		\Sigma_{(U,V)|W}=\Sigma_{U,V}-\Sigma_{U,W}\Sigma^{-1}_{W}\Sigma_{U,W}^T
		\end{equation*}
	\end{itemize}
	\end{Proposition}
\textbf{Proof of Proposition \ref{propo:optfairEOpred}.} 
In the particular normal model, this independency means that the elements in positions $(1,2)$ and $(2,1)$ of the covariance matrix of random vector $(g(X,S), S \mid Y)$ are exactly zero. Therefore, the class of fair predictors is written as
\begin{equation}\label{eq:EOcondition}
\mathcal{F}_{EO}:=\{g:\mathcal{X}\times\mathcal{S}\rightarrow R^d: Cov(g(X,S), S \mid Y)=0\}
\end{equation}
More precisely, previous condition can be written in terms of the covariances of $(X,S,Y)$ and the coefficients $(\beta_0,\beta)$ of the linear model \eqref{equ:linearnormalmodel}. Observe that the joint distribution of the random vector $(g_{\beta_{0},\beta}(X,S),S,Y)$ is
\begin{equation*}
\left[\begin{array}{c}
\beta_0S+ \beta^TX\\
S\\
Y
\end{array}\right]\sim \mathcal{N} \left(\left[\begin{array}{c}
\beta_0\mu_S+\beta^T\mu_X\\
\mu_S\\
\mu_Y
\end{array}\right], \left[\begin{array}{cc}
\Sigma_1& \Sigma_{12}\\
\Sigma_{12}^T& \Sigma_Y 
\end{array}\right] \right),
\end{equation*}
where
\begin{align*}
&\Sigma_1=\left[\begin{array}{cc}
\beta_0^2\Sigma_S+\beta^T\Sigma_X\beta+2\beta_0\beta^T\Sigma_{XS} & \beta_0\Sigma_S+\beta^T\Sigma_{XS} \\
\beta_0\Sigma_{SY}+\beta^T\Sigma_{XY}& \Sigma_S \\
\end{array}\right] \in \mathbb{R}^{2 \times 2},\\
&\Sigma_{12}=\left[\begin{array}{c}
\beta_0\Sigma_{SY}+\beta^T\Sigma_{XY}\\
\Sigma_{SY}
\end{array}\right] \in \mathbb{R}^{2 \times 1}.
\end{align*}
Hence, from Proposition \ref{prop:gaussian}, we know that $$Cov(g(X,S), S \mid Y)=\Sigma_1-\frac{1}{\Sigma_Y}\Sigma_{12}\Sigma_{12}^T.$$
Substituting the expressions above for $\Sigma_1$ and $\Sigma_2$, we obtain that $g_{\beta_{0},\beta} \in \mathcal{F}_{EO} $ if and only if
\begin{align*}
(\beta_0\Sigma_S+\beta^T\Sigma_{XS})\Sigma_Y=\Sigma_{SY}(\beta_0\Sigma_{SY}+\beta^T\Sigma_{XY}).
\end{align*}
Then the optimal EO-fair predictor in this setting is the solution to the following optimization problem:
\begin{align}\label{optimalEOfair_NR}
&\left(\hat{\beta}_{0,fair},\hat{\beta}_{fair}\right):=\argmin_{(\beta_0,\beta)\in \mathcal{F}_{EO}} \mathbb{E}\left[\left(Y-g_{\beta_0,\beta}(X,S)\right)^2\right]\\
\mathcal{F}_{EO}&=\{(\beta_0,\beta)\in \R\times\R^p \ \text{such that} \ \beta^T\left(\Sigma_{XS}\Sigma_Y-\Sigma_{SY}\Sigma_{XY}\right)+\beta_0\left(\Sigma_S\Sigma_Y-\Sigma_{SY}^2\right)=0\} \nonumber.
\end{align}
We note that Cauchy-Schwarz inequality together with the assumption that $Y$ and $S$ are not linearly dependent ensure $\Sigma_S\Sigma_Y-\Sigma_{SY}^2 > 0$. Then we obtain that the class of EO-fair predictors $(\beta_0,\beta)\in \mathcal{F}_{EO}$ are such that $\beta_0=\beta^TC_{S,X,Y},$ where
\begin{equation*}
C_{S,X,Y}:=\left(\frac{\Sigma_{XS}\Sigma_Y-\Sigma_{SY}\Sigma_{XY}}{\Sigma_S\Sigma_Y-\Sigma_{SY}^2}\right) \in \mathbb{R}^{p\times 1}.
\end{equation*}
Hence, the optimal EO-fair predictor \eqref{optimalEOfair_NR} can be obtained equivalently
\begin{equation*}
\hat{\beta}_{fair}=\argmin_{\beta \in \R^{p}}\mathbb{E}\left[\left(Y-\beta^T(X+SC_{S,X,Y})\right)^2\right].
\end{equation*}
Now if we denote $Z:=X+SC_{S,X,Y}$, it is easy to check that the optimal EO-fair predictor can be exactly computed as
\begin{align*}\label{optEOpredictor}
\hat{\beta}_{fair}&=\Sigma_Z^{-1}\Sigma_{Z,Y}, \ \text{where}\\
\Sigma_Z&=\Sigma_X+\Sigma_SC_{S,X,Y}C_{S,X,Y}^T+C_{S,X,Y}\Sigma_{XS}^T+\Sigma_{XS}C_{S,X,Y}^T \nonumber\\
\Sigma_{ZY}&=\Sigma_{XY}+\Sigma_{SY}C_{S,X,Y} \nonumber.
\end{align*}

$\hfill\Box$\\

\subsection{Proofs of section \ref{subsubsec:PFEOclass}} \label{app:optfairEOclass}
\textbf{Proof of Proposition \ref{propo:optEOclass}.} Let us consider the following minimization problem
\begin{equation*}
(*):=\min_{g \in \mathcal{G}}\{\mathcal{R}(g): \mathbb{P}(g(X,S)=i \mid Y=i, S=1)=\mathbb{P}(g(X,S)=i \mid Y=i, S=0), i=0,1\}.
\end{equation*}
Using the weak duality we can write
\begin{align*}
(*)&=\\
=&\min_{g \in \mathcal{G}}\max_{(\lambda_0, \lambda_1) \in \R^2}\left\lbrace\mathcal{R}(g)+ \sum_{i=0,1}\lambda_i\left[\mathbb{P}(g(X,S)=i \mid Y=i, S=1)-\mathbb{P}(g(X,S)=i \mid Y=i, S=0)\right]\right\rbrace\\
\geq& \max_{(\lambda_0, \lambda_1) \in \R^2}\min_{g \in \mathcal{G}}\left\lbrace\mathcal{R}(g)+ \sum_{i=0,1}\lambda_i\left[\mathbb{P}(g(X,S)=i \mid Y=i, S=1)-\mathbb{P}(g(X,S)=i \mid Y=i, S=0)\right]\right\rbrace\\
=:&(**).
\end{align*}
We first study the objective function of the max min problem $(**)$, which is equal to
\begin{align*}
\mathbb{P}(g(X,S)\neq Y)+ \sum_{i=0,1}\lambda_i\left(\mathbb{P}(g(X,S)=i \mid Y=i, S=1)-\mathbb{P}(g(X,S)=i \mid Y=i, S=0)\right).
\end{align*}
The first step of the proof is to simplify the expression above to linear functional of the classifier $g$. Notice that we can write for the first term
\begin{align*}
\mathbb{P}(g(X,S)\neq Y)= &\mathbb{P}(g(X,S)=0, Y=1)+\mathbb{P}(g(X,S)=1, Y=0)\\
=&\mathbb{P}(g(X,S)=1) + \mathbb{P}(Y=1)-\mathbb{P}(g(X,S)=1, Y=1)-\mathbb{P}(g(X,S)=1, Y=1)\\
=&\mathbb{P}(g(X,S)=1) + \mathbb{P}(Y=1)-2\mathbb{P}(g(X,S)=1, Y=1)\\
=&\mathbb{P}(Y=1)+\mathbb{E}\left[g(X,S)\right]-2\mathbb{P}(S=1)\mathbb{E}\left[\mathbbm{1}_{g(X,S)=1,Y=1} \mid S=1\right]\\
&-2\mathbb{P}(S=0)\mathbb{E}\left[\mathbbm{1}_{g(X,S)=1,Y=1} \mid S=0\right]\\
=&\mathbb{P}(Y=1)-\mathbb{P}(S=1)\mathbb{E}_{X \mid S=1}\left[g(X,1)(2 \eta(X,1)-1)\right]\\
&-\mathbb{P}(S=0)\mathbb{E}_{X \mid S=0}\left[g(X,0)(2 \eta(X,0)-1)\right].
\end{align*}

Moreover, for $s=0,1,$ we can write for the rest four terms in the objetive function
\begin{align*}
&\mathbb{P}(g(X,S)=1 \mid Y=1, S=s)=\frac{\mathbb{P}(g(X,S)=1 , Y=1 \mid S=s)}{\mathbb{P}(Y=1 \mid S=s)}=\frac{\mathbb{E}_{X \mid S=s}\left[g(X,s)\eta(X,s)\right]}{\mathbb{P}(Y=1 \mid S=s)}\\
&\mathbb{P}(g(X,S)=0 \mid Y=0, S=s)=1-\mathbb{P}(g(X,S)=1 \mid Y=0, S=s)\\
&=1-\frac{\mathbb{E}_{X \mid S=s}\left[g(X,s)(1-\eta(X,s))\right]}{\mathbb{P}(Y=0 \mid S=s)}.
\end{align*}
Using these, the objective of $(**)$ can be simplified as
\begin{align*}
\mathbb{P}(Y=1)+\mathbb{E}_{X \mid S=1}\Big[\Big.g(X,1)&\left(\eta(X,1)\left(\frac{\lambda_1}{\mathbb{P}(Y=1\mid S=1)}+\frac{\lambda_0}{1-\mathbb{P}(Y=1\mid S=1)}-2\mathbb{P}(S=1)\right)\right. \\
& \left.\left.+\mathbb{P}(S=1)-\frac{\lambda_0}{1-\mathbb{P}(Y=1\mid S=1)}\right)\right]\\
+\mathbb{E}_{X \mid S=0}\Big[\Big.g(X,0)&\left(\eta(X,0)\left(-\frac{\lambda_1}{\mathbb{P}(Y=1\mid S=0)}-\frac{\lambda_0}{1-\mathbb{P}(Y=1\mid S=0)}-2\mathbb{P}(S=0)\right)\right.\\
& \left.\left.+\mathbb{P}(S=0)+\frac{\lambda_0}{1-\mathbb{P}(Y=1\mid S=0)}\right)\right].
\end{align*}
For every $\mathbf{\lambda}:=(\lambda_0, \lambda_1) \in \R^2$ a minimizer $g_{\lambda}^*$ of the problem $(**)$ can be written for all $x \in \R^d$ as
\begin{align*}
g_{\lambda}^*(x,1)&=\mathbbm{1}_{\{\eta(X,1)\left(\frac{\lambda_1}{\mathbb{P}(Y=1\mid S=1)}+\frac{\lambda_0}{1-\mathbb{P}(Y=1\mid S=1)}-2\mathbb{P}(S=1)\right)
	+\mathbb{P}(S=1)-\frac{\lambda_0}{1-\mathbb{P}(Y=1\mid S=1)} \leq 0\}} \\
&=\mathbbm{1}_{\{1-\eta(X,1)\left(2-\frac{\lambda_1}{\mathbb{P}(Y=1, S=1)}-\frac{\lambda_0}{\mathbb{P}(Y=0, S=1)}\right)-\frac{\lambda_0}{\mathbb{P}(Y=0, S=1)} \leq 0\}} \\
&=\mathbbm{1}_{\{1\leq 2\eta(X,1)-\lambda_1\frac{\eta(X,1)}{\mathbb{P}(Y=1, S=1)}+\lambda_0\frac{1-\eta(X,1)}{\mathbb{P}(Y=0, S=1)}\}}\\
g_{\lambda}^*(x,0)&=\mathbbm{1}_{\{\eta(X,0)\left(-\frac{\lambda_1}{\mathbb{P}(Y=1\mid S=0)}-\frac{\lambda_0}{1-\mathbb{P}(Y=1\mid S=0)}-2\mathbb{P}(S=0)\right)
	+\mathbb{P}(S=0)+\frac{\lambda_0}{1-\mathbb{P}(Y=1\mid S=0)}\leq 0\}}\\
&=\mathbbm{1}_{\{1-\eta(X,0)\left(2+\frac{\lambda_1}{\mathbb{P}(Y=1, S=0)}+\frac{\lambda_0}{\mathbb{P}(Y=0, S=0)}\right)+\frac{\lambda_0}{\mathbb{P}(Y=0, S=0)} \leq 0\}}\\
&=\mathbbm{1}_{\{1\leq 2\eta(X,0)+\lambda_1\frac{\eta(X,0)}{\mathbb{P}(Y=1, S=0)}-\lambda_0\frac{1-\eta(X,0)}{\mathbb{P}(Y=0, S=0)}\}}.
\end{align*}

It is interesting to observe that for $\lambda_0=0$ we recover the optimal equal opportunity classifier obtained first in \cite{chzhen2019leveraging}. If in addition  $\lambda_1=0$, then we recover the Bayes classifier. Now, substituting this classifier into the objective of $(**)$ we arrive at
\begin{align*}
\mathbb{P}(Y=1)-\min_{(\lambda_0,\lambda_1)\in \R^2}\Big\{ \mathbb{E}_{X \mid S=1}&\Big[\eta(X,1)\left(-2\mathbb{P}(S=1)+\frac{\lambda_1}{\mathbb{P}(Y=1\mid S=1)}+\frac{\lambda_0}{1-\mathbb{P}(Y=1\mid S=1)}\right) \\
& +\mathbb{P}(S=1)-\frac{\lambda_0}{1-\mathbb{P}(Y=1\mid S=1)}\Big]_+\\
+\mathbb{E}_{X \mid S=0}&\Big[\eta(X,0)\left(-2\mathbb{P}(S=0)-\frac{\lambda_1}{\mathbb{P}(Y=1\mid S=0)}-\frac{\lambda_0}{1-\mathbb{P}(Y=1\mid S=0)}\right)\\
& +\mathbb{P}(S=0)+\frac{\lambda_0}{1-\mathbb{P}(Y=1\mid S=0)}\Big]_+\Big\}.
\end{align*}
We observe that the mappings
\begin{align*}
(\lambda_0,\lambda_1)\mapsto \mathbb{E}_{X \mid S=1}&\Big[\eta(X,1)\left(-2\mathbb{P}(S=1)+\frac{\lambda_1}{\mathbb{P}(Y=1\mid S=1)}+\frac{\lambda_0}{1-\mathbb{P}(Y=1\mid S=1)}\right) \\
& +\mathbb{P}(S=1)-\frac{\lambda_0}{1-\mathbb{P}(Y=1\mid S=1)}\Big]_+\\
(\lambda_0,\lambda_1)\mapsto  \mathbb{E}_{X \mid S=0}&\Big[\eta(X,0)\left(-2\mathbb{P}(S=0)-\frac{\lambda_1}{\mathbb{P}(Y=1\mid S=0)}-\frac{\lambda_0}{1-\mathbb{P}(Y=1\mid S=0)}\right)\\
& +\mathbb{P}(S=0)+\frac{\lambda_0}{1-\mathbb{P}(Y=1\mid S=0)}\Big]_+
\end{align*}
are convex, therefore we can write the first order optimality conditions as
\begin{align*}
\mathbf{0} \in \partial_{\lambda}\mathbb{E}_{X \mid S=1}&\Big[\eta(X,1)\left(-2\mathbb{P}(S=1)+\frac{\lambda_1}{\mathbb{P}(Y=1\mid S=1)}-\frac{\lambda_0}{1-\mathbb{P}(Y=1\mid S=1)}\right) \\
& +\mathbb{P}(S=1)-\frac{\lambda_0}{1-\mathbb{P}(Y=1\mid S=1)}\Big]_+\\
+\partial_{\lambda}\mathbb{E}_{X \mid S=0}&\Big[\eta(X,0)\left(-2\mathbb{P}(S=0)-\frac{\lambda_1}{\mathbb{P}(Y=1\mid S=0)}-\frac{\lambda_0}{1-\mathbb{P}(Y=1\mid S=0)}\right)\\
& +\mathbb{P}(S=0)+\frac{\lambda_0}{1-\mathbb{P}(Y=1\mid S=0)}\Big]_+
\end{align*}
Under Assumption \ref{ass:optEOclass} this subgradient is reduced to the gradient almost surely, thus we have the following two conditions on the optimal value of $\lambda^*$
\begin{align}
\frac{\mathbb{E}_{X \mid S=1}\Big[\eta(X,1)g_{\lambda^*}^*(X,1)\Big]}{\mathbb{P}(Y=1\mid S=1)}&=\frac{\mathbb{E}_{X \mid S=0}\Big[\eta(X,0)g_{\lambda^*}^*(X,0)\Big]}{\mathbb{P}(Y=1\mid S=0)} \label{cond1:optEO}\\
\frac{\mathbb{E}_{X \mid S=1}\Big[(1-\eta(X,1))g_{\lambda^*}^*(X,1)\Big]}{\mathbb{P}(Y=0\mid S=1)}&=\frac{\mathbb{E}_{X \mid S=0}\Big[(1-\eta(X,0))g_{\lambda^*}^*(X,0)\Big]}{\mathbb{P}(Y=0\mid S=0)}\label{cond2:optEO}
\end{align}
and the pair $(\lambda^*,g_{\lambda^*}^*)$ is a solution of the dual problem $(**)$. 
By the definition of the regression function \eqref{def:regressionfunction}, we note that previous conditions \eqref{cond1:optEO} and \eqref{cond2:optEO} can be written as
\begin{align*}
\mathbb{P}(g_{\lambda^*}^*(X,1)=1\mid Y=1,S=1)&=\mathbb{P}(g_{\lambda^*}^*(X,0)=1\mid Y=1,S=0)\\
\mathbb{P}(g_{\lambda^*}^*(X,1)=1\mid Y=0,S=1)&=\mathbb{P}(g_{\lambda^*}^*(X,1)=1\mid Y=0,S=0)
\end{align*}
which implies that the classifier $g_{\lambda^*}^* \in \mathcal{F}_{EO}$, that is, it is fair in the EO sense.

Finally, it remains to show that $g_{\lambda^*}^*$ is actually an optimal classifier. Indeed, since $g_{\lambda^*}^*$ is fair we can write on the one hand
\begin{align*}
\mathcal{R}(g_{\lambda^*}^*)\geq \min_{g \in \mathcal{G}}\{\mathcal{R}(g): \mathbb{P}(g(X,S)=i \mid Y=i, S=0)=\mathbb{P}(g(X,S)=i \mid Y=i, S=1), i=0,1\}=(*).
\end{align*}
On the other hand, the pair $(\lambda^*,g_{\lambda^*}^*)$ is a solution of the dual problem $(**)$, thus we have
\begin{align*}
(*) &\geq \mathcal{R}(g_{\lambda^*}^*)+ \sum_{i=0,1}\lambda^*_i\left(\mathbb{P}(g_{\lambda^*}^*(X,S)=i \mid Y=i, S=0)-\mathbb{P}(g_{\lambda^*}^*(X,S)=i \mid Y=i, S=1)\right)\}\\ &=\mathcal{R}(g_{\lambda^*}^*).
\end{align*}
It implies that the classifier $g_{\lambda^*}^*$ is optimal, hence $g^*\equiv g_{\lambda^*}^*$.

$\hfill \Box$\\


\begin{thebibliography}{10}
	
	\bibitem{adler2018auditing}
	P.~Adler, C.~Falk, S.~A Friedler, T.~Nix, G.~Rybeck, C.~Scheidegger, B.~Smith,
	and S.~Venkatasubramanian.
	\newblock Auditing black-box models for indirect influence.
	\newblock {\em Knowledge and Information Systems}, 54(1):95--122, 2018.
	
	\bibitem{agarwal2018reductions}
	A.~Agarwal, A.~Beygelzimer, M.~Dud{\'\i}k, J.~Langford, and H.~Wallach.
	\newblock A reductions approach to fair classification.
	\newblock {\em arXiv preprint arXiv:1803.02453}, 2018.
	
	\bibitem{agarwal2019fair}
	A.~Agarwal, M.~Dud{\'\i}k, and Z.~S. Wu.
	\newblock Fair regression: Quantitative definitions and reduction-based
	algorithms.
	\newblock {\em arXiv preprint arXiv:1905.12843}, 2019.
	
	\bibitem{agueh2011barycenters}
	M.~Agueh and G.~Carlier.
	\newblock Barycenters in the wasserstein space.
	\newblock {\em SIAM Journal on Mathematical Analysis}, 43(2):904--924, 2011.
	
	\bibitem{ali2019loss}
	J.~Ali, M.~B. Zafar, A.~Singla, and K.~P. Gummadi.
	\newblock Loss-aversively fair classification.
	\newblock In {\em Proceedings of the 2019 AAAI/ACM Conference on AI, Ethics,
		and Society}, pages 211--218, 2019.
	
	\bibitem{barocas2016big}
	S.~Barocas and A.~D. Selbst.
	\newblock Big data's disparate impact.
	\newblock {\em Calif. L. Rev.}, 104:671, 2016.
	
	\bibitem{barocas-hardt-narayanan}
	Solon Barocas, Moritz Hardt, and Arvind Narayanan.
	\newblock {\em Fairness and Machine Learning}.
	\newblock fairmlbook.org, 2019.
	\newblock \url{http://www.fairmlbook.org}.
	
	\bibitem{bechavod2017penalizing}
	Y.~Bechavod and K.~Ligett.
	\newblock Penalizing unfairness in binary classification.
	\newblock {\em arXiv preprint arXiv:1707.00044}, 2017.
	
	\bibitem{belkin2006manifold}
	M.~Belkin, P.~Niyogi, and V.~Sindhwani.
	\newblock Manifold regularization: A geometric framework for learning from
	labeled and unlabeled examples.
	\newblock {\em Journal of machine learning research}, 7(Nov):2399--2434, 2006.
	
	\bibitem{berk2017convex}
	R.~Berk, H.~Heidari, S.~Jabbari, M.~Joseph, M.~Kearns, J.~Morgenstern, S.~Neel,
	and A.~Roth.
	\newblock A convex framework for fair regression.
	\newblock {\em arXiv preprint arXiv:1706.02409}, 2017.
	
	\bibitem{berk2018fairness}
	R.~Berk, H.~Heidari, S.~Jabbari, M.~Kearns, and A.~Roth.
	\newblock Fairness in criminal justice risk assessments: The state of the art.
	\newblock {\em Sociological Methods \& Research}, page 0049124118782533, 2018.
	
	\bibitem{beutel2017data}
	A.~Beutel, J.~Chen, Z.~Zhao, and E.~H. Chi.
	\newblock Data decisions and theoretical implications when adversarially
	learning fair representations.
	\newblock {\em arXiv preprint arXiv:1707.00075}, 2017.
	
	\bibitem{black2020fliptest}
	E~Black, S~Yeom, and M~Fredrikson.
	\newblock Fliptest: fairness testing via optimal transport.
	\newblock In {\em Proceedings of the 2020 Conference on Fairness,
		Accountability, and Transparency}, pages 111--121, 2020.
	
	\bibitem{bousquet2004introduction}
	O.~Bousquet, S.~Boucheron, and G.~Lugosi.
	\newblock Introduction to statistical learning theory.
	\newblock In {\em Advanced lectures on machine learning}, pages 169--207.
	Springer, 2004.
	
	\bibitem{calders2009building}
	T.~Calders, F.~Kamiran, and M.~Pechenizkiy.
	\newblock Building classifiers with independency constraints.
	\newblock In {\em 2009 IEEE International Conference on Data Mining Workshops},
	pages 13--18. IEEE, 2009.
	
	\bibitem{calders2010three}
	T.~Calders and S.~Verwer.
	\newblock Three naive bayes approaches for discrimination-free classification.
	\newblock {\em Data Mining and Knowledge Discovery}, 21(2):277--292, 2010.
	
	\bibitem{calmon2017optimized}
	F.~Calmon, D.~Wei, B.~Vinzamuri, K.~N. Ramamurthy, and K.~R. Varshney.
	\newblock Optimized pre-processing for discrimination prevention.
	\newblock In {\em Advances in Neural Information Processing Systems}, pages
	3992--4001, 2017.
	
	\bibitem{chiappa2020general}
	S.~Chiappa, R.~Jiang, T.~Stepleton, A.~Pacchiano, H.~Jiang, and J.~Aslanides.
	\newblock A general approach to fairness with optimal transport.
	\newblock In {\em Thirty-Fourth AAAI Conference on Artificial Intelligence},
	2020.
	
	\bibitem{chierichetti2017fair}
	F.~Chierichetti, S.~Kumar, R.and~Lattanzi, and S.~Vassilvitskii.
	\newblock Fair clustering through fairlets.
	\newblock In {\em Advances in Neural Information Processing Systems}, pages
	5029--5037, 2017.
	
	\bibitem{chouldechova2017fair}
	A.~Chouldechova.
	\newblock Fair prediction with disparate impact: A study of bias in recidivism
	prediction instruments.
	\newblock {\em Big data}, 5(2):153--163, 2017.
	
	\bibitem{chzhen2019leveraging}
	E.~Chzhen, C.~Denis, M.~Hebiri, L.~Oneto, and M.~Pontil.
	\newblock Leveraging labeled and unlabeled data for consistent fair binary
	classification.
	\newblock In {\em Advances in Neural Information Processing Systems}, pages
	12739--12750, 2019.
	
	\bibitem{corbett2017algorithmic}
	S.~Corbett-Davies, E.~Pierson, A.~Feller, S.~Goel, and A.~Huq.
	\newblock Algorithmic decision making and the cost of fairness.
	\newblock In {\em Proceedings of the 23rd ACM SIGKDD International Conference
		on Knowledge Discovery and Data Mining}, pages 797--806. ACM, 2017.
	
	\bibitem{cotter2018training}
	A.~Cotter, M.~Gupta, H.~Jiang, N.~Srebro, K.~Sridharan, S.~Wang, B.~Woodworth,
	and S.~You.
	\newblock Training well-generalizing classifiers for fairness metrics and other
	data-dependent constraints.
	\newblock {\em arXiv preprint arXiv:1807.00028}, 2018.
	
	\bibitem{dauphin2014identifying}
	Y.~N. Dauphin, R.~Pascanu, C.~Gulcehre, K.~Cho, S.~Ganguli, and Y.~Bengio.
	\newblock Identifying and attacking the saddle point problem in
	high-dimensional non-convex optimization.
	\newblock In {\em Advances in neural information processing systems}, pages
	2933--2941, 2014.
	
	\bibitem{del2019central}
	E.~Del~Barrio and J-M. Loubes.
	\newblock Central limit theorems for empirical transportation cost in general
	dimension.
	\newblock {\em The Annals of Probability}, 47(2):926--951, 2019.
	
	\bibitem{dieterich2016compas}
	W.~Dieterich, C.~Mendoza, and T.~Brennan.
	\newblock Compas risk scales: Demonstrating accuracy equity and predictive
	parity.
	\newblock {\em Northpoint Inc}, 2016.
	
	\bibitem{doherty2012information}
	Neil~A Doherty, Anastasia~V Kartasheva, and Richard~D Phillips.
	\newblock Information effect of entry into credit ratings market: The case of
	insurers' ratings.
	\newblock {\em Journal of Financial Economics}, 106(2):308--330, 2012.
	
	\bibitem{donini2018empirical}
	M.~Donini, L.~Oneto, S.~Ben-David, J.~S. Shawe-Taylor, and M.~Pontil.
	\newblock Empirical risk minimization under fairness constraints.
	\newblock In {\em Advances in Neural Information Processing Systems}, pages
	2791--2801, 2018.
	
	\bibitem{dwork2012fairness}
	C.~Dwork, M.~Hardt, T.~Pitassi, O.~Reingold, and R.~Zemel.
	\newblock Fairness through awareness.
	\newblock In {\em Proceedings of the 3rd innovations in theoretical computer
		science conference}, pages 214--226. ACM, 2012.
	
	\bibitem{dwork2018decoupled}
	C.~Dwork, N.~Immorlica, A.~T. Kalai, and M.~Leiserson.
	\newblock Decoupled classifiers for group-fair and efficient machine learning.
	\newblock In {\em Conference on Fairness, Accountability and Transparency},
	pages 119--133, 2018.
	
	\bibitem{edwards2015censoring}
	H.~Edwards and A.~Storkey.
	\newblock Censoring representations with an adversary.
	\newblock In {\em 4th International Conference on Learning Representations},
	2015.
	
	\bibitem{edwards2016towards}
	H.~Edwards and A.~Storkey.
	\newblock Towards a neural statistician.
	\newblock {\em arXiv preprint arXiv:1606.02185}, 2016.
	
	\bibitem{feldman2015computational}
	Michael Feldman.
	\newblock {\em Computational Fairness: Preventing Machine-Learned
		Discrimination}.
	\newblock PhD thesis, 2015.
	
	\bibitem{FFMSV}
	S.~A Feldman, M.and~Friedler, J.~Moeller, C.~Scheidegger, and
	S.~Venkatasubramanian.
	\newblock Certifying and removing disparate impact.
	\newblock In {\em Proceedings of the 21th ACM SIGKDD International Conference
		on Knowledge Discovery and Data Mining}, pages 259--268. ACM, 2015.
	
	\bibitem{fish2016confidence}
	B.~Fish, J.~Kun, and {\'A}.~D. Lelkes.
	\newblock A confidence-based approach for balancing fairness and accuracy.
	\newblock In {\em Proceedings of the 2016 SIAM International Conference on Data
		Mining}, pages 144--152. SIAM, 2016.
	
	\bibitem{fish2015fair}
	J.~Fish, B.and~Kun and A.~D. Lelkes.
	\newblock Fair boosting: a case study.
	\newblock In {\em Workshop on Fairness, Accountability, and Transparency in
		Machine Learning}. Citeseer, 2015.
	
	\bibitem{flores2016false}
	A.~W. Flores, K.~Bechtel, and C.T. Lowenkamp.
	\newblock False positives, false negatives, and false analyses: A rejoinder to
	machine bias: There's software used across the country to predict future
	criminals. and it's biased against blacks.
	\newblock {\em Fed. Probation}, 80:38, 2016.
	
	\bibitem{fukuchi2015prediction}
	K.~Fukuchi, T.~Kamishima, and J.~Sakuma.
	\newblock Prediction with model-based neutrality.
	\newblock {\em IEICE TRANSACTIONS on Information and Systems},
	98(8):1503--1516, 2015.
	
	\bibitem{gano2017disparate}
	A.~Gano.
	\newblock Disparate impact and mortgage lending: A beginner's guide.
	\newblock {\em U. Colo. L. Rev.}, 88:1109, 2017.
	
	\bibitem{ghassami2018fairness}
	A.~Ghassami, S.~Khodadadian, and N.~Kiyavash.
	\newblock Fairness in supervised learning: An information theoretic approach.
	\newblock In {\em 2018 IEEE International Symposium on Information Theory
		(ISIT)}, pages 176--180. IEEE, 2018.
	
	\bibitem{goh2016satisfying}
	G.~Goh, A.~Cotter, M.~Gupta, and M.~P. Friedlander.
	\newblock Satisfying real-world goals with dataset constraints.
	\newblock In {\em Advances in Neural Information Processing Systems}, pages
	2415--2423, 2016.
	
	\bibitem{gordaliza2019obtaining}
	P.~Gordaliza, E.~Del~Barrio, F.~Gamboa, and J.-M. Loubes.
	\newblock Obtaining fairness using optimal transport theory.
	\newblock In {\em International Conference on Machine Learning}, pages
	2357--2365, 2019.
	
	\bibitem{gretton2005kernel}
	A.~Gretton, R.~Herbrich, A.~Smola, O.~Bousquet, and B.~Sch{\"o}lkopf.
	\newblock Kernel methods for measuring independence.
	\newblock {\em Journal of Machine Learning Research}, 6(Dec):2075--2129, 2005.
	
	\bibitem{DBLP:journals/corr/abs-1712-07924}
	P.~Hacker and E.~Wiedemann.
	\newblock A continuous framework for fairness.
	\newblock {\em CoRR}, abs/1712.07924, 2017.
	
	\bibitem{hajian2012injecting}
	S.~Hajian, A.~Monreale, D.~Pedreschi, J.~Domingo-Ferrer, and F.~Giannotti.
	\newblock Injecting discrimination and privacy awareness into pattern
	discovery.
	\newblock In {\em 2012 IEEE 12th International Conference on Data Mining
		Workshops}, pages 360--369. IEEE, 2012.
	
	\bibitem{hardt2016equality}
	M.~Hardt, E.~Price, and N.~Srebro.
	\newblock Equality of opportunity in supervised learning.
	\newblock In {\em Advances in neural information processing systems}, pages
	3315--3323, 2016.
	
	\bibitem{hebert2018calibration}
	U.~H{\'e}bert-Johnson, M.~P. Kim, O.~Reingold, and G.~N. Rothblum.
	\newblock Calibration for the (computationally-identifiable) masses.
	\newblock In {\em International Conference on Machine Learning}, pages
	1939--1948, 2018.
	
	\bibitem{jiang2019wasserstein}
	R.~Jiang, A.~Pacchiano, T.~Stepleton, H.~Jiang, and S.~Chiappa.
	\newblock Wasserstein fair classification.
	\newblock In {\em Thirty-Fifth Uncertainty in Artificial Intelligence
		Conference}, 2019.
	
	\bibitem{johndrow2019algorithm}
	J.~E. Johndrow and K.~Lum.
	\newblock An algorithm for removing sensitive information: application to
	race-independent recidivism prediction.
	\newblock {\em The Annals of Applied Statistics}, 13(1):189--220, 2019.
	
	\bibitem{kamiran2009classifying}
	F.~Kamiran and T.~Calders.
	\newblock Classifying without discriminating.
	\newblock In {\em 2009 2nd International Conference on Computer, Control and
		Communication}, pages 1--6. IEEE, 2009.
	
	\bibitem{kamiran2010classification}
	F.~Kamiran and T.~Calders.
	\newblock Classification with no discrimination by preferential sampling.
	\newblock In {\em Proc. 19th Machine Learning Conf. Belgium and The
		Netherlands}, pages 1--6. Citeseer, 2010.
	
	\bibitem{kamiran2012data}
	F.~Kamiran and T.~Calders.
	\newblock Data preprocessing techniques for classification without
	discrimination.
	\newblock {\em Knowledge and Information Systems}, 33(1):1--33, 2012.
	
	\bibitem{kamiran2012decision}
	F.~Kamiran, A.~Karim, and X.~Zhang.
	\newblock Decision theory for discrimination-aware classification.
	\newblock In {\em 2012 IEEE 12th International Conference on Data Mining},
	pages 924--929. IEEE, 2012.
	
	\bibitem{kamishima2012fairness}
	T.~Kamishima, S.~Akaho, H.~Asoh, and J.~Sakuma.
	\newblock Fairness-aware classifier with prejudice remover regularizer.
	\newblock In {\em Joint European Conference on Machine Learning and Knowledge
		Discovery in Databases}, pages 35--50. Springer, 2012.
	
	\bibitem{kamishima2011fairness}
	T.~Kamishima, S.~Akaho, and J.~Sakuma.
	\newblock Fairness-aware learning through regularization approach.
	\newblock In {\em 2011 IEEE 11th International Conference on Data Mining
		Workshops}, pages 643--650. IEEE, 2011.
	
	\bibitem{kearns2018preventing}
	M.~Kearns, S.~Neel, A.~Roth, and Z.~S. Wu.
	\newblock Preventing fairness gerrymandering: Auditing and learning for
	subgroup fairness.
	\newblock In {\em International Conference on Machine Learning}, pages
	2564--2572, 2018.
	
	\bibitem{kilbertus2017avoiding}
	N.~Kilbertus, M.~Rojas-Carulla, G.~Parascandolo, M.~Hardt, D.~Janzing, and
	B.~Sch{\"o}lkopf.
	\newblock Avoiding discrimination through causal reasoning.
	\newblock In {\em Advances in Neural Information Processing Systems}, pages
	656--666, 2017.
	
	\bibitem{kim2019multiaccuracy}
	M.~P. Kim, A.~Ghorbani, and J.~Zou.
	\newblock Multiaccuracy: Black-box post-processing for fairness in
	classification.
	\newblock In {\em Proceedings of the 2019 AAAI/ACM Conference on AI, Ethics,
		and Society}, pages 247--254, 2019.
	
	\bibitem{kleinberg2016inherent}
	J.~Kleinberg, S.~Mullainathan, and M.~Raghavan.
	\newblock Inherent trade-offs in the fair determination of risk scores.
	\newblock {\em arXiv preprint arXiv:1609.05807}, 2016.
	
	\bibitem{komiyama2018nonconvex}
	A.~Komiyama, J.and~Takeda, J.~Honda, and H.~Shimao.
	\newblock Nonconvex optimization for regression with fairness constraints.
	\newblock In {\em International conference on machine learning}, pages
	2737--2746, 2018.
	
	\bibitem{le2017existence}
	T.~Le~Gouic and J-M. Loubes.
	\newblock Existence and consistency of wasserstein barycenters.
	\newblock {\em Probability Theory and Related Fields}, 168(3-4):901--917, 2017.
	
	\bibitem{legouic2020fair}
	T.~Le~Gouic and J-M. Loubes.
	\newblock Fair regression in $l^2$: optimal fair prediction for demographic
	parity.
	\newblock {\em arXiv preprint arXiv:3192298}, 2020.
	
	\bibitem{li2019kernel}
	Z.~Li, A.~Perez-Suay, G.~Camps-Valls, and D.~Sejdinovic.
	\newblock Kernel dependence regularizers and gaussian processes with
	applications to algorithmic fairness.
	\newblock {\em arXiv preprint arXiv:1911.04322}, 2019.
	
	\bibitem{loftus2018causal}
	J~R Loftus, C~Russell, M~J Kusner, and R~Silva.
	\newblock Causal reasoning for algorithmic fairness.
	\newblock {\em arXiv preprint arXiv:1805.05859}, 2018.
	
	\bibitem{madras2018learning}
	D.~Madras, E.~Creager, T.~Pitassi, and R.~Zemel.
	\newblock Learning adversarially fair and transferable representations.
	\newblock {\em arXiv preprint arXiv:1802.06309}, 2018.
	
	\bibitem{madras2018predict}
	D.~Madras, T.~Pitassi, and R.~Zemel.
	\newblock Predict responsibly: improving fairness and accuracy by learning to
	defer.
	\newblock In {\em Advances in Neural Information Processing Systems}, pages
	6147--6157, 2018.
	
	\bibitem{mary2019fairness}
	J.~Mary, C.~Calauzenes, and N.~El~Karoui.
	\newblock Fairness-aware learning for continuous attributes and treatments.
	\newblock In {\em International Conference on Machine Learning}, pages
	4382--4391, 2019.
	
	\bibitem{mehrabi2019survey}
	N.~Mehrabi, F.~Morstatter, N.~Saxena, K.~Lerman, and A.~Galstyan.
	\newblock A survey on bias and fairness in machine learning.
	\newblock {\em arXiv preprint arXiv:1908.09635}, 2019.
	
	\bibitem{menon2018cost}
	A.~K. Menon and R.~C. Williamson.
	\newblock The cost of fairness in binary classification.
	\newblock In {\em Conference on Fairness, Accountability and Transparency},
	pages 107--118, 2018.
	
	\bibitem{mercat2016discrimination}
	M.~Mercat-Bruns.
	\newblock {\em Discrimination at Work}.
	\newblock University of California Press, 2016.
	
	\bibitem{kusner2017counterfactual}
	Kusner M.J., Loftus J., Russell C., and Silva R.
	\newblock Counterfactual fairness.
	\newblock In {\em Advances in Neural Information Processing Systems}, pages
	4066--4076, 2017.
	
	\bibitem{nabi2018fair}
	R.~Nabi and I.~Shpitser.
	\newblock Fair inference on outcomes.
	\newblock In {\em Thirty-Second AAAI Conference on Artificial Intelligence},
	2018.
	
	\bibitem{narasimhan2018learning}
	H.~Narasimhan.
	\newblock Learning with complex loss functions and constraints.
	\newblock In {\em International Conference on Artificial Intelligence and
		Statistics}, pages 1646--1654, 2018.
	
	\bibitem{noriega2019active}
	A.~Noriega-Campero, M.~A. Bakker, B.~Garcia-Bulle, and A.~Pentland.
	\newblock Active fairness in algorithmic decision making.
	\newblock In {\em Proceedings of the 2019 AAAI/ACM Conference on AI, Ethics,
		and Society}, pages 77--83, 2019.
	
	\bibitem{oneto2019taking}
	L.~Oneto, M.~Donini, A.~Elders, and M.~Pontil.
	\newblock Taking advantage of multitask learning for fair classification.
	\newblock In {\em Proceedings of the 2019 AAAI/ACM Conference on AI, Ethics,
		and Society}, pages 227--237, 2019.
	
	\bibitem{oneto2020recent}
	Navarin N. Sperduti~A. Oneto, L. and D.~Anguita.
	\newblock {\em Recent Trends in Learning from Data}.
	\newblock Springer, 2020.
	
	\bibitem{pedreschi2009measuring}
	D.~Pedreschi, S.~Ruggieri, and F.~Turini.
	\newblock Measuring discrimination in socially-sensitive decision records.
	\newblock In {\em Proceedings of the 2009 SIAM international conference on data
		mining}, pages 581--592. SIAM, 2009.
	
	\bibitem{pedreshi2008discrimination}
	D.~Pedreshi, S.~Ruggieri, and F.~Turini.
	\newblock Discrimination-aware data mining.
	\newblock In {\em Proceedings of the 14th ACM SIGKDD international conference
		on Knowledge discovery and data mining}, pages 560--568, 2008.
	
	\bibitem{perez2017fair}
	A.~P{\'e}rez-Suay, V.~Laparra, G.~Mateo-Garc{\'\i}a, J.~Mu{\~n}oz-Mar{\'\i},
	L.~G{\'o}mez-Chova, and G.~Camps-Valls.
	\newblock Fair kernel learning.
	\newblock In {\em Joint European Conference on Machine Learning and Knowledge
		Discovery in Databases}, pages 339--355. Springer, 2017.
	
	\bibitem{quadrianto2017recycling}
	N.~Quadrianto and V.~Sharmanska.
	\newblock Recycling privileged learning and distribution matching for fairness.
	\newblock In {\em Advances in Neural Information Processing Systems}, pages
	677--688, 2017.
	
	\bibitem{ribeiro2016should}
	M.~T. Ribeiro, S.~Singh, and C.~Guestrin.
	\newblock " why should i trust you?" explaining the predictions of any
	classifier.
	\newblock In {\em Proceedings of the 22nd ACM SIGKDD international conference
		on knowledge discovery and data mining}, pages 1135--1144, 2016.
	
	\bibitem{risser2019using}
	L.~Risser, Q.~Vincenot, N.~Couellan, and J-M. Loubes.
	\newblock Using wasserstein-2 regularization to ensure fair decisions with
	neural-network classifiers.
	\newblock {\em arXiv preprint arXiv:1908.05783}, 2019.
	
	\bibitem{serrurier2019fairness}
	M.~Serrurier, J-M. Loubes, and E.~Pauwels.
	\newblock Fairness with wasserstein adversarial networks.
	\newblock Technical report, working paper or preprint, 2019.
	
	\bibitem{siegel2014race}
	R.~B. Siegel.
	\newblock Race-conscious but race-neutral: The constitutionality of disparate
	impact in the roberts court.
	\newblock {\em Ala. L. Rev.}, 66:653, 2014.
	
	\bibitem{simoiu2017problem}
	C.~Simoiu, S.~Corbett-Davies, S.~Goel, et~al.
	\newblock The problem of infra-marginality in outcome tests for discrimination.
	\newblock {\em The Annals of Applied Statistics}, 11(3):1193--1216, 2017.
	
	\bibitem{song2018learning}
	J.~Song, P.~Kalluri, A.~Grover, S.~Zhao, and S.~Ermon.
	\newblock Learning controllable fair representations.
	\newblock {\em arXiv preprint arXiv:1812.04218}, 2018.
	
	\bibitem{speicher2018unified}
	T.~Speicher, H.~Heidari, N.~Grgic-Hlaca, K.~P. Gummadi, A.~Singla, and M.~B.
	Weller, A.and~Zafar.
	\newblock A unified approach to quantifying algorithmic unfairness: Measuring
	individual \&group unfairness via inequality indices.
	\newblock In {\em Proceedings of the 24th ACM SIGKDD International Conference
		on Knowledge Discovery \& Data Mining}, pages 2239--2248, 2018.
	
	\bibitem{steinberg2020fast}
	D.~Steinberg, A.~Reid, S.~O'Callaghan, F.~Lattimore, L.~McCalman, and
	T.~Caetano.
	\newblock Fast fair regression via efficient approximations of mutual
	information.
	\newblock {\em arXiv preprint arXiv:2002.06200}, 2020.
	
	\bibitem{verma2018fairness}
	S.~Verma and J.~Rubin.
	\newblock Fairness definitions explained.
	\newblock In {\em 2018 IEEE/ACM International Workshop on Software Fairness
		(FairWare)}, pages 1--7. IEEE, 2018.
	
	\bibitem{villani2008optimal}
	C.~Villani.
	\newblock {\em Optimal transport: old and new}, volume 338.
	\newblock Springer Science \& Business Media, 2008.
	
	\bibitem{woodworth2017learning}
	B.~Woodworth, S.~Gunasekar, M.~I. Ohannessian, and N.~Srebro.
	\newblock Learning non-discriminatory predictors.
	\newblock {\em arXiv preprint arXiv:1702.06081}, 2017.
	
	\bibitem{yona2018probably}
	G.~Yona and G.~Rothblum.
	\newblock Probably approximately metric-fair learning.
	\newblock In {\em International Conference on Machine Learning}, pages
	5680--5688, 2018.
	
	\bibitem{ZVGG}
	M~B Zafar, I~Valera, M~Gomez~Rodriguez, and K~P Gummadi.
	\newblock Fairness beyond disparate treatment \& disparate impact: Learning
	classification without disparate mistreatment.
	\newblock In {\em Proceedings of the 26th International Conference on World
		Wide Web}, pages 1171--1180. International World Wide Web Conferences
	Steering Committee, 2017.
	
	\bibitem{zafar2019fairness}
	M.~B. Zafar, I.~Valera, M.~Gomez-Rodriguez, and K.~P. Gummadi.
	\newblock Fairness constraints: A flexible approach for fair classification.
	\newblock {\em Journal of Machine Learning Research}, 20(75):1--42, 2019.
	
	\bibitem{zafar2017parity}
	M.~B. Zafar, I.~Valera, M.~Rodriguez, K.~Gummadi, and A.~Weller.
	\newblock From parity to preference-based notions of fairness in
	classification.
	\newblock In {\em Advances in Neural Information Processing Systems}, pages
	229--239, 2017.
	
	\bibitem{zehlike2020matching}
	M.~Zehlike, P.~Hacker, and E.~Wiedemann.
	\newblock Matching code and law: achieving algorithmic fairness with optimal
	transport.
	\newblock {\em Data Mining and Knowledge Discovery}, 34(1):163--200, 2020.
	
	\bibitem{zemel2013learning}
	R.~Zemel, Y.~Wu, K.~Swersky, T.~Pitassi, and C.~Dwork.
	\newblock Learning fair representations.
	\newblock In {\em International Conference on Machine Learning}, pages
	325--333, 2013.
	
	\bibitem{zhang2018mitigating}
	B.~H. Zhang, B.~Lemoine, and M.~Mitchell.
	\newblock Mitigating unwanted biases with adversarial learning.
	\newblock In {\em Proceedings of the 2018 AAAI/ACM Conference on AI, Ethics,
		and Society}, pages 335--340, 2018.
	
	\bibitem{zliobaite2015relation}
	I.~Zliobaite.
	\newblock On the relation between accuracy and fairness in binary
	classification.
	\newblock {\em arXiv preprint arXiv:1505.05723}, 2015.
	
\end{thebibliography}

\end{document}